\documentclass[graybox]{svmult}

\usepackage{type1cm}        
\usepackage{makeidx}         
\usepackage{graphicx}        
\usepackage{multicol}        
\usepackage[bottom]{footmisc}

\usepackage{newtxtext}       %
\usepackage[varvw]{newtxmath}       
\usepackage{natbib}
\usepackage{hyperref}
\usepackage{cprotect}
\usepackage{multirow}

\makeindex             

\newcommand{\eg}{\textit{e.g. }}
\newcommand{\ie}{\textit{i.e. }}
\newcommand{\etc}{\textit{etc}}

\AtEndPreamble{
    \usepackage[capitalize]{cleveref}
    \crefname{section}{Sec.}{Secs.}
    \Crefname{section}{Section}{Sections}
    \Crefname{table}{Table}{Tables}
    \crefname{table}{Tab.}{Tabs.}
}

\begin{document}

\title*{\centerline{Recent Deep Learning in Crowd Behaviour Analysis: A Brief Review}
}

\author{\centerline{Jiangbei Yue and He Wang*}\hfill\break\\
\centerline{$\copyright$ 2025, Springer Nature}\hfill\break
\centerline{All rights reserved.}\\[19pt]
\centerline{January, 2025}
}
\institute{Jiangbei Yue \at School of Computer Science, University of Leeds, Leeds LS2 9JT, United Kingdom.\\ \\
He Wang, corresponding author \at UCL Centre for Artificial Intelligence, Department of Computer Science, University College London, London NW1 2AE, United Kingdom \\ \email{he\_wang@ucl.ac.uk}}

\maketitle

\abstract{Crowd behaviour analysis is essential to numerous real-world applications, such as public safety and urban planning, and therefore has been studied for decades. In the last decade or so, the development of deep learning has significantly propelled the research on crowd behaviours. This chapter reviews recent advances in crowd behaviour analysis using deep learning. We mainly review the research in two core tasks in this field, crowd behaviour prediction and recognition. We broadly cover how different deep neural networks, after first being proposed in machine learning, are applied to analysing crowd behaviours. This includes pure deep neural network models as well as recent development of methodologies combining physics with deep learning. In addition, representative studies are discussed and compared in detail. Finally, we discuss the effectiveness of existing methods and future research directions in this rapidly evolving field. This chapter aims to provide a high-level summary of the ongoing deep learning research in crowd behaviour analysis. It intends to help new researchers who just entered this field to obtain an overall understanding of the ongoing research, as well as to provide a retrospective analysis for existing researchers to identify possible future directions.}

\section{Introduction}
The research in crowd behaviour analysis aims to understand, recognise, and predict crowds in different situations, environments, events, \etc. The behaviour arises from the aggregation of a group of people sharing the same physical space~\citep{wijermans2011understanding,swathi2017crowd}. It is a highly interdisciplinary and cross-disciplinary field, where a wide range of research topics have been investigated, from individual body movements to the overall crowd flow, from the low-level physical constraints to the high-level socio-psychological factors~\citep{zhan2008crowd,kok2016crowd,murino2017group}. Given its wide impact on a wide range of applications, \eg safety, security, event organisation, and transportation, physicists, mathematicians, computer scientists, and psychologists have jointly explored this area for decades.

In this chapter, we focus on recent research in crowd behaviour analysis in computer science, specifically in deep learning. The emergence of deep learning has advanced the research in Artificial Intelligence (AI) to its ever-greatest today. As a universal computational machinery, deep neural networks have been employed to improve or even replace many research tools in many fields, including crowd behaviour analysis. Given the fast pace of the development of deep learning, it is particularly important to conduct a timely review of the most recent deep learning based research in this field, to summarise the up-to-date achievements, compare different approaches, and discuss the possible future directions of how crowd behaviour analysis can continue to leverage the cutting-edge AI technologies.

The application of deep learning in any field requires two key components: data and task definition. The former is the foundation of AI while the latter enables specific methods to be developed and metrics to be chosen for evaluation. As the history of deep learning in crowd behaviour analysis is relatively short, it covers only a small fraction of the problems that have been widely studied in other fields. This is probably mostly due to the lack of (big) data and probably also the unfamiliarity of the modelling tools to people outside of computer science and engineering. Currently, one of the first and most active research communities for crowd behaviour analysis is computer vision. It is not surprising that computer vision has become an active research field for understanding crowd behaviours. Video data is easy to obtain, and cameras as sensors are commonly accepted. In fact, cameras are not only often used but are the only sensors used in many places, especially in public and communal spaces, e.g. CCTV cameras, because of their non-invasiveness. Therefore, large amounts of video data have been obtained for crowds, which establishes the foundation for deep learning for crowd behaviour analysis. 

In addition to abundant data, computer vision is also one of the first fields where crowd behaviour analysis tasks are defined for deep learning. So far, there are mainly two tasks: behaviour prediction and recognition. Therefore, in this chapter, we focus on reviewing the most recent research in these two subfields. Specifically, crowd behaviour prediction involves forecasting the future states or actions of a crowd, with or without details on the individuals in it, based on their observed behaviours~\citep{fan2015citymomentum,alahi2016social,jiang2021deepcrowd}. This task emphasises modelling the evolving patterns in crowd movements and is essential for applications such as crowd management and autonomous vehicles. Crowd behaviour recognition, on the other hand, aims to identify the type of behaviours exhibited by a given crowd, focusing on analysing the patterns in crowds, which are often regarded as a whole~\citep{khokher2014crowd}. Such recognition can be applied in domains such as public safety and automatic surveillance~\citep{khokher2014crowd,qaraqe2024crowd}. 

Besides the tasks, there are still other factors to consider when classifying the current research in a review. An important factor is the type of model. In the past decade, new types of deep neural networks have been proposed one after another at a fast pace. Since there are unique advantages and disadvantages for each type of new network, it is unclear which type of network or what combination of them would be the best for crowd analysis. Therefore, we have seen wave after wave of research applying the latest deep learning model for crowd analysis. Therefore, we will also use the types of neural networks as one major factor to classify recent research in both tasks. Next, given the fast-growing number of papers being published in this field, it is impractical to dive into the details of each one of them. So we only give details of the methodology of some of the representative papers. Finally, since some of the deep learning research in this field is closely related to the traditional data-driven approaches, \ie statistical machine learning, which also has been active in this field, we also briefly review these methods which are from approximately the same period of time.

\section{Crowd Behaviour Prediction}
Crowd behaviours can be studied from many perspectives at different levels~\citep{thida2013literature,yang2020review}. One perspective is microscopic which focuses on individual-level movements, where various levels of details have been modelled, from simplifying individuals into 2D discs to modelling detailed 3D body motions~\citep{van2008reciprocal,feldmann2023forward,gomez2024resolving}. After modelling the individuals, the collective motions are regarded as crowd behaviours~\citep{yang2020review}. In contrast with the microscopic level, there is also a macroscopic perspective. This type of research often treats the crowd as a whole \eg as a continuum~\citep{hughes2002continuum,golas2014continuum}. 

Microscopic analysis is currently trendy in computer vision. One popular modelling perspective is to treat each individual as a 2D disc moving in a 2D environment. The required data are extracted from crowd videos and are normally in the form of trajectories, often in 2D but also occasionally in 3D. These trajectories are either human-labelled or automatically estimated via tracking algorithms~\citep{pellegrini2009you,robicquet2016learning}. Trajectory data has been extensively used to understand individual behaviours and group flows in traditional statistical machine learning~\citep{wang2016path,he2020informative}, as well as recent deep learning~\citep{sighencea2021review,gu2022stochastic,lin2024progressive}. One core evaluation protocol for these methods is to see whether a model can actually predict the short-horizon and long-horizon future. This is evaluated by several metrics which are based on the difference between the predicted trajectories by the trained model and the ground-truth trajectories. This difference can be measured by trajectory-wise difference, \eg accuracy, or distributions of trajectories, \eg distributional divergences.

\subsection{Traditional Machine Learning Methods}
To predict trajectories, traditional machine learning methods typically rely heavily on feature engineering for a good representation of the crowd/individual state and their dynamics. Popular models include Gaussian processes~\citep{mackay1998introduction}, Markov models~\citep{fosler1998markov}, Kalman filters~\citep{bishop2001introduction}, \etc. \citet{ellis2009modelling} explicitly modelled the probabilistic distribution of the current velocities based on a Gaussian process prior. Given the current positions, instantaneous velocities are then sampled from the estimated distributions to infer the next positions. This process is performed recursively to predict the future trajectories. \citet{kitani2012activity} propose a unified model based on hidden variable Markov decision processes. With the prior knowledge of people's goals, they employed semantic scene labelling and inverse optimal control to model the people-people interaction and the people-environment interaction. BRVO~\citep{kim2015brvo} follows the reciprocal velocity obstacle~\citep{van2008reciprocal} formulation to build a prediction model, where the ensemble Kalman filter~\citep{evensen2003ensemble} is exploited to optimise the relevant parameters. Although these methods require data to calibrate model parameters, the amount of data needed is significantly less than what is required for deep learning methods. 

Besides individual trajectory prediction, statistical machine learning methods have also been employed to predict trajectories with similar distributions, \ie not hinged on individual motions but collective group motions. However, this requires analysing the flow patterns of trajectories and developing relevant metrics. As labelling flows is too laborious to be practical for crowd trajectories, these methods are often based on unsupervised learning and have been developed to cluster trajectories to identify flows~\citep{wang2011trajectory,wang2016path,wang2017trending,wang2016globally,he2020informative}. Aiming to find natural clusters of similar behaviours in crowds, these methods automatically discover spatial similarities~\citep{wang2011trajectory,wang2016path,wang2017trending}, spatial-temporal similarities~\citep{wang2016globally}, or spatial-temporal-dynamics similarities~\citep{he2020informative}. Along with such analysis, new metrics have been proposed to measure whether prediction/simulation is similar to real crowds. These metrics are based on rules~\citep{singh2009steerbench}, or overall statistical similarities~\citep{guy2012statistical}, scene semantics of flows~\citep{wang2016path,wang2017trending}, \etc., so that prediction and simulation can be guided by these metrics~\citep{he2020informative}.

Overall, there is prolific research in using traditional statistical machine learning for crowd behaviour prediction. In comparison with the recent deep learning methods reviewed below, these methods tend to require fewer data samples, employ more white-box models, and have better explainability of the prediction. However, when compared based on prediction accuracy, they are generally inferior to deep learning.

\subsection{Deep Learning Methods}

Deep learning methods have recently dominated the field of human trajectory prediction with exceptional prediction accuracy. Since neural networks are universal function approximators~\citep{hornik1989multilayer,hornik1991approximation}, they can learn complex and non-linear dynamics and patterns from data, especially when learning from a large amount of data. Despite being almost black-box, it significantly reduces the reliance on handcrafted features, hence less reliance on the expertise and time required for manual feature engineering. At the same time, the techniques for data collection, \eg tracking algorithms in video analysis, have been fast developing, resulting in abundant data. As a result, this has significantly extended the basis of researchers on pedestrian and crowd analysis, especially in the field of computer vision and AI. 

Following the development of deep learning itself, a broad set of powerful neural networks, each with its unique characteristics, have been proposed and adapted for human trajectory prediction. These models include Recurrent Neural Networks (RNNs)~\citep{fausett1994fundamentals}, Convolutional Neural Networks (CNNs)~\citep{lecun1989handwritten}, Graph Neural Networks (GNNs)~\citep{wu2020comprehensive}, generative models~\citep{goodfellow2014generative,sohn2015learning,sohl2015deep}, transformers~\citep{vaswani2017attention}, \etc. Researchers focus on how to leverage diverse network architectures to capture motion dynamics. Therefore, we can broadly classify existing deep learning methods into RNN-based, CNN-based, GNN-based, generative model-based, and transformer-based methods according to the network architecture. Since this research community has been extremely prolific in the past decade, we will cover only some representative papers.


\subsubsection{Recurrent Neural Networks} 
RNNs are a class of neural networks with neurons with an internal state. They can model sequential data with dependence by connecting their output with their input and changing the internal state accordingly, which captures the non-linear dependence between data at different time steps or different positions in a sequence. Specifically, given a sequence, RNNs utilize the same network repeatedly to process each element in the sequence, where the output for the current step is used as a part of the input at the next step. This design enables RNNs to capture temporal dependencies, which is vital for tasks such as human trajectory prediction. 

Inspired by the success of RNNs in other sequence prediction tasks \eg speech generation, \citet{alahi2016social} first applied RNNs in trajectory prediction and achieved superior prediction accuracy, which is the pioneering work in trajectory prediction based on deep learning. The authors proposed a new model called Social-LSTM, based on the Long Short Term Memory (LSTM) network~\citep{hochreiter1997long} (a classic RNN). One technical and modelling novelty in this method is a module which embeds a learnable component in the network to capture the interactions between pedestrians. With each trajectory captured by one LSTM for each pedestrian, a novel social pooling layer is introduced to model the interactions between pedestrians, by connecting the features of the neighbouring pedestrians learned by their LSTMs. The output of the social pooling layer at each step is fed back to the LSTM of the pedestrian in interest as part of the input for the next step. Since Social-LSTM marks the beginning of the new deep learning chapter in human trajectory prediction, a more detailed review of the method is below. 

\begin{figure}[tb]
    \centering
    \includegraphics[width=0.95\linewidth]{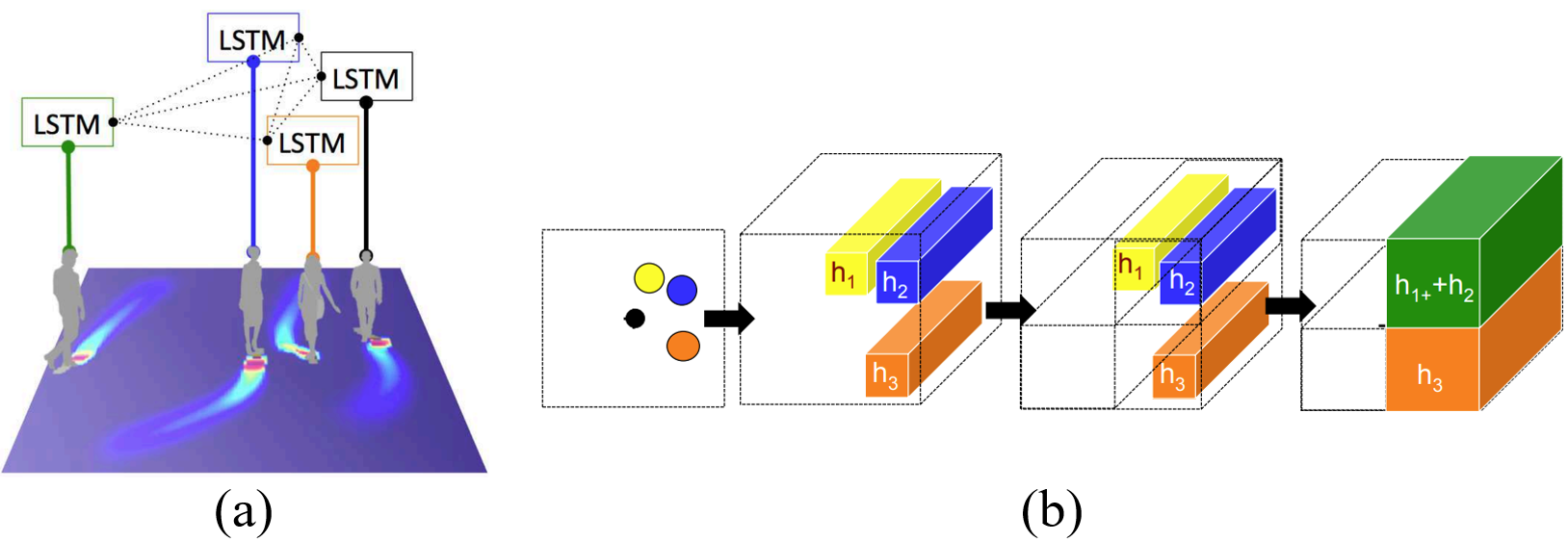}
    \caption{Overview of the Social-LSTM (a) and visualization of constructing social tensors (b). The black dot and other dots (yellow, blue and orange) denote the target person and neighbours, respectively, in (b). These two figures are from~\citet{alahi2016social}.}
    \label{fig:social-lstm}
\end{figure}

As shown in \cref{fig:social-lstm} (a), one LSTM is employed for each person to model their dynamics, where all LSTMs share the same weights. The LSTM takes as input the current location and interaction features to update its internal states. Specifically, the current location of the i$th$ person $(x_t^i, y_t^i)$ at the time step $t$ is transformed into the embedding feature $e^i_t$ by a multi-layer perceptron (MLP). Meanwhile, the social pooling builds a grid near the person in interest, as shown in \cref{fig:social-lstm} (b), and constructs a $N_o \times N_o \times D$ social tensor $H_t^i$:
\begin{align}
    H_t^i(m,n,:) = \sum_{j \in N_i} \textbf{1}_{mn}[x^j_t-x^i_t, y^j_t - y^i_t]h_{t-1}^j,
\end{align}
where $N_o$ is the neighbourhood size, D is the dimension of the hidden states $h_t$ of the LSTM, $\textbf{1}_{mn}[x,y]$ is an indicator function showing if $(x,y)$ is located in the $(m,n)$ cell of the grid, $N_i$ denotes the set containing neighbours of the i$th$ person. Whether a pedestrian is considered a neighbour of the pedestrian in interest is based on their distance at the current time. When the distance between the i$th$ and the j$th$ person falls below a pre-defined threshold, the j$th$ person is considered a neighbour of the i$th$ person. Note the construction of social tensors preserves the spatial information. 

Next, the social tensor $H_t^i$ is then fed into an MLP to obtain the corresponding embedding feature $a_t^i$ (\cref{eq:emb_pooling}). Subsequently, the LSTM takes as input the last hidden state $h_{t-1}^i$ and embedded features ($e_t^i, a_t^i$) to output the new hidden state (\cref{eq:lstm_update}). Overall, the whole process is formulated as the following recurrence:
\begin{align}
    \label{eq:emb_location} e_t^i &= MLP_e(x_t^i, y_t^i),  \\
    \label{eq:emb_pooling} a_t^i &= MLP_a(H_t^i), \\ 
    \label{eq:lstm_update} h_t^i &= LSTM(h_{t-1}^{i}, e_t^i, a_t^i),
\end{align}
where the initial hidden state $h_0$ is given. $e_t^i$ is a feature of location $(x_t^i, y_t^i)$.

Lastly, the hidden states $h_t^i$ are used to estimate the distribution of the next position $(x_{t+1}^i, y_{t+1}^i)$. Social-LSTM assumes that the next position follows a bivariate Gaussian distribution, which has five parameters: the mean $\mu_{t+1}^i=(\mu_x, \mu_y)_{t+1}^i$, the standard deviation $\sigma_{t+1}^i=(\sigma_x, \sigma_y)_{t+1}^i$, and the correlation coefficient $\rho_{t+1}^i$. The hidden state $h_t^i$ is fed into an MLP to predict the distribution parameters of the next position:
\begin{align}
    [\mu_{t+1}^i, \sigma_{t+1}^i, \rho_{t+1}^i] = MLP_p(h_t^i).
\end{align}
Finally, one can sample $(\hat{x}_{t+1}^i, \hat{y}_{t+1}^i)$ from the predicted distribution $\mathcal{N}(\mu_{t+1}^i, \sigma_{t+1}^i, \rho_{t+1}^i)$. 

Since Social-LSTM explicitly parameterises the distribution of the next position as a Gaussian, assuming conditional independence between different time steps and between different pedestrians, one can easily write down the joint likelihood of all observed pedestrian positions as a product of the likelihood of individual positions, given their respective Gaussian parameters. Therefore, Social-LSTM optimises the model parameters by minimising the negative log-likelihood:
\begin{align}
    L = - \sum_i \sum_{t=T_h+1}^{T_h+T_f} \log(P(x_t^i, y_t^i | \mu_{t}^i, \sigma_{t}^i, \rho_{t}^i)),
\end{align}
where the first summation involves all pedestrians, $T_h$ is the length of the history trajectory, and $T_f$ is the length of the future trajectory. Overall, one of the key novelties in Social-LSTM is the learnable interactions between pedestrians embedded into the design of the RNN. Starting from Social-LSTM, a line of RNN-based methods has been proposed.

\citet{xue2018ss} proposed a hierarchical LSTM network named Social-Scene-LSTM based on an encoder-decoder framework, to incorporate the environment factor which is ignored by Social-LSTM. Social-Scene-LSTM uses three LSTM encoders for three scales (person, social, and scene) and an LSTM decoder to estimate future trajectories. The model considers the observed individual trajectories, the positions of the neighbours, and the frames of the video at every time step. These features correspond to the single-person, the social, and the scene scales, respectively. The LSTM encoder outputs the corresponding features at different scales. These features are concatenated and fed into the LSTM decoder to obtain the predicted trajectories. Later, in contrast to the Social-Scene-LSTM which considers the influence of the environment by simply encoding the frames, context-aware LSTM~\citep{bartoli2018context} followed the framework of Social-LSTM and proposed a novel context-aware pooling layer to model the interactions between people and objects in the environment. \citet{zhang2021pedestrian} used Gated Recurrent Units (GRU)~\citep{chung2014empirical} (also an RNN) to replace the LSTM in Social-LSTM to avoid overfitting. They also emphasized that the encoded representation of the trajectory positions is more effective than the raw positions when serving as the input to the GRUs. Given the importance of destinations in trajectory prediction,~\citet{tran2021goal} proposed a dual-channel neural network, consisting of a goal channel and a trajectory channel, based on GRUs. Specifically, the goal-channel network employs GRUs to extract features from the estimates of multiple possible destinations. The model converts the extracted features into a series of control signals by a flexible attention mechanism. Finally, the trajectory channel network predicts the future trajectories under the guidance of the control signals.   

In addition to the methods reviewed above, there are abundant other RNN-based methods with various improvements. There is a series of methods which improve the performance of the social pooling mechanism by considering the motion coherence of groups \citep{bisagno2018group,bisagno2021embedding}, modelling the importance or attention to neighbours~\citep{fernando2018soft+,xu2018encoding,shi2019pedestrian}, adding the repulsion constraint~\citep{xu2018collision}, \etc. Moreover, \citet{zhang2019sr} designed a state refinement module based on a message-passing framework to enhance the representation of hidden states of LSTMs and optimize interactions. \citet{zhu2019starnet} proposed the StartNet with a star topology to model interactions between individuals. \citet{liang2019peeking} presented a multi-task learning framework, modelling predictions of activities and trajectories jointly.

\subsubsection{Convolutional Neural Networks} 
CNN (Convolutional Neural Network) is a type of feed-forward neural network with convolution layers, which excels in capturing the correlations between the features of a location with its neighbours. This type of network has been employed to learn from data with strong temporal and spatial correlations and provide high computational efficiency, compared with \eg multilayer perceptron networks. Therefore, many methods incorporate CNNs to predict pedestrian trajectories. 

The key to applying CNNs is to organise the data or features so that the concept of vicinity can be defined. As one of the earliest methods applying CNNs for trajectory prediction, \citet{nikhil2018convolutional} encoded each position of the past trajectory to obtain the embedding features and organise these features in an image-like form. Then, a CNN is used to apply convolutions on these features and outputs a global feature across all the historical positions. Lastly, the global feature is fed into a fully connected layer to predict future trajectories. 

\citet{zamboni2022pedestrian} extended the CNN model in~\citet{nikhil2018convolutional} by building different mappings between the global feature and each of the frames in the future trajectory. This is done by dividing the output of the CNN in \citep{nikhil2018convolutional} into N features instead of a single global feature, where N is the number of time steps in the future trajectory. These N features are separately fed into a fully connected layer to predict their corresponding future positions. In addition, the authors explored the effectiveness of 1D convolution, 2D convolution, positional embeddings, transpose convolutional layers, and residual connections in CNNs for trajectory prediction. 

Next, to further improve the modelling of the interactions between pedestrians, Social-IWSTCNN \citep{zhang2021social} introduces a new social interaction extractor module based on MLPs and an aggregate function to extract features containing social and spatial information from observed trajectories. Subsequently, a temporal convolutional network is proposed to encode the interaction features for refinement by further incorporating the temporal information. The refined features are fed into another CNN to predict distributions of future trajectories. 

\citet{zhang2022step} proposed to combine CNNs with RNNs to improve the performance in trajectory prediction. Specifically, they used two LSTMs to encode the input trajectory first to extract the time-related features for each time step. Then, every feature is reshaped and fed as input into a CNN to mine high-dimensional patterns within it. The outputs of the CNN are further enhanced based on an augmented attention mechanism to capture the global information. Then a GRU followed by an MLP takes the enhanced features as the input to predict the next position. The proposed model recursively estimates the future trajectory. 

Beyond the papers above, the idea of combining CNNs with RNNs is also utilised by many other people~\citep{ridel2020scene,song2020pedestrian,de2021social,shafiee2021introvert,jain2020discrete}.  

\subsubsection{Graph Neural Networks} 
GNNs are a class of neural networks specifically designed to process data structured as graphs defined by a set of nodes and a set of edges. They have a wide range of applications where the data does not lie in the Euclidean space. Originally, a trajectory is naturally regarded as 2D Euclidean data. However, a group of trajectories from different pedestrians as a whole can be modelled differently. The introduction of GNNs brings an alternative view of them. Trajectories from pedestrians together can be regarded as a graph, \eg if each pedestrian is viewed as a node and the edges between nodes are used to model the interactions between pedestrians. This way, GNNs have the ability to effectively model the spatial, temporal, and relational dynamics in complex graphs~\citep{yan2018spatial,wu2020comprehensive}. This has motivated various GNN-based methods for trajectory prediction. 

\citet{huang2019stgat} adopted graph attention networks~\citep{velivckovic2018graph} and LSTMs to extract information from spatio-temporal graphs constructed based on past trajectories and predict future motions. Social-STGCNN~\citep{mohamed2020social} constructs spatio-temporal graphs $G$ consisting of a set of spatial graphs $G_t=\{V_t, E_t\}$ in all the past time steps to represent the historical trajectories, where nodes in $V_t$ denote pedestrians, edges in $E_t$ connect a pedestrian with their neighbours. Each node in $V_t$ has attributes such as locations and each edge in $E_t$ is associated with weights describing the interaction intensity \eg whether two pedestrians are close so that they need to steer clear of each other. Subsequently, the proposed spatio-temporal graph convolution neural network takes the spatio-temporal graphs as input to extract the spatio-temporal features. Finally, time-extrapolator CNNs are used to predict the future trajectories based on the extracted features.

The concept of spatio-temporal graphs is further employed in many other methods~\citep{wang2021graphtcn,zhou2021ast,lian2023ptp,sighencea2023d}. DAG-Net~\citep{monti2021dag} uses two graph neural networks to estimate the destinations of pedestrians and model the interactions between people. Then a variational RNN~\citep{chung2015recurrent} is employed to incorporate the information of the estimated destinations and the interactions, to obtain the future trajectories by recursively predicting the next position. \citet{shi2021sgcn} argued that previous research typically models redundant interactions between pedestrians. To address the problem, they proposed a sparse graph convolution network model. Specifically, the model introduces a new sparse graph learning based on the self-attention mechanism~\citep{vaswani2017attention}. It also uses asymmetric convolution networks to construct sparse spacial graphs with sparse directed interactions, and sparse temporal graphs with motion tendencies, from the input trajectories. Subsequently, graph convolution networks~\citep{kipf2016semi} are used to extract the features from these sparse graphs. The learned features are then fed into a time convolution network to estimate the future trajectories.        

\subsubsection{Generative Models} 
Generative models are machine learning models that can generate similar data to the data on which they are trained. These models normally achieve this by learning a mapping between the unknown data distribution and some simpler distribution where the sampling is easier, \eg a Gaussian or uniform distribution. In deep learning, these methods include Generative Adversarial Networks (GANs)~\citep{goodfellow2014generative}, Variational Autoencoders (VAEs)~\citep{kingma2013auto}, diffusion models~\citep{sohl2015deep}, \etc. One advantage of generative models is that they are suitable for modelling randomness in non-deterministic processes, which makes them useful in trajectory prediction. This is because there is intrinsic randomness in pedestrian trajectories due to the influence of various unobserved factors \eg environment and affective states~\citep{luo2008agent,itkina2023interpretable}. Therefore, the mapping from some historical trajectory to the future trajectory can be seen as a stochastic process where generative models can be employed. 

GANs consist of a generator network and a discriminator network. The generator is trained to generate synthetic samples resembling real samples, while the discriminator is trained to tell the difference between synthetic and real samples. The adversarial training is for the generator to be able to fool the discriminator, and for the discriminator to be able to distinguish synthetic samples from the real ones, until an equilibrium is reached. Social-GAN~\citep{gupta2018social} is one of the earliest research papers which applies GANs to trajectory prediction. In Social-GAN, the generator is an LSTM-based encoder-decoder network. The discriminator is simply an LSTM-based encoder followed by a classification layer. Given the past trajectories and the noises drawn from a distribution, the generator is trained to predict future trajectories. Additionally, Social-GAN designs a novel pooling mechanism applied to the generator to model the interactions between individuals. It also employs a new adversarial loss function to obtain diverse predictions which are consistent with the observed trajectories. The initial success of Social-GAN has inspired other similar research. SoPhie~\citep{sadeghian2019sophie} is a GAN-based method incorporating the restriction of the environment \eg walls, obstacles. The model first leverages CNNs to extract the environment features from the video frames and employs LSTMs to extract the movement features for each person. Then, an attention module is used to capture the physical attention and the social attention, to optimize the environment features and the motion features, respectively. Then both features are concatenated for each person. These concatenated features are finally fed into an LSTM-based GAN to estimate the distributions of the future trajectories. In addition, other methods also introduce different variants of GANs to model crowd dynamics. \citet{amirian2019social} employed Info-GAN~\citep{chen2016infogan} to avoid mode collapsing and dropping in naive GANs while \citet{kosaraju2019social} utilised Bicycle-GAN~\citep{zhu2017toward} to improve the modelling of multi-modal trajectories.

Other than GANs, another type of generative model, VAEs, has also led to a stream of new methods for trajectory prediction. The key idea of VAEs is to assume there is a latent space where the distribution of data is a Gaussian, so that an encode-decoder network can be trained where the encoder maps the data into the latent space and the decoder reconstructs the data from the latent space. After training, samples can be drawn from the Gaussian and then decoded into outputs which are similar to the training data. Often, the encoder and decoder also take as input an additional condition, where it becomes a Conditional VAE~\citep{sohn2015learning}, or CVAE. PECNet~\citep{mangalam2020not} uses a CVAE to predict the distributions of destinations based on the observed trajectories, which is further used to guide the prediction of future trajectories through a new social pooling mechanism. Here, the condition of the CVAE is the past trajectory. \citet{lee2022muse} also employed a CVAE to estimate destinations but another CVAE was used to generate full trajectories. SocialVAE~\citep{xu2022socialvae} combines a CVAE with a RNN. In this work, the encoder and decoder in the CVAE are implemented as two GRUs to reconstruct the future trajectories directly, while another GRU extracts the features from the historical trajectories to serve as the condition of the CVAE. \citet{yu2024pedestrian} modified SocialVAE by using complex gated recurrent units~\citep{xu2023s} as the encoder and decoder in the CVAE and a graph attention network to extract condition features. \citet{zhou2022dynamic} combined CVAEs with GANs to predict trajectories. It adopts the architecture of GANs, with a CVAE-based generator and an RNN-based discriminator.

Recently, diffusion models have gained significant interest in many domains such as image generation, due to their strong generative capabilities. Diffusion models generate data by first adding noises to the data so that they can be seen as samples from a Gaussian. Then the model learns to denoise these noise-polluted samples from the Gaussian to reconstruct the data samples. Once learned, a mapping between a Gaussian and the data distribution is established. Again, this positions diffusion models as natural candidates for modelling stochasticity, which has inspired researchers to use diffusion models in stochastic trajectory prediction. \citet{gu2022stochastic} is one of the first to do so.  In~\citet{gu2022stochastic}, the denoising process is constructed from all walkable areas to the desired trajectory distributions within some limited areas. Additionally, the past trajectories are encoded into state embeddings serving as conditions of the denoising process. Although the model in~\citet{gu2022stochastic} can balance the diversity and the accuracy of prediction by controlling the denoising steps, they typically need a large number of denoising steps, which is time-consuming and hinders its application. To address the issue, \citet{mao2023leapfrog} proposed a new diffusion-based model with a leapfrog initialiser. Instead of denoising from all walkable areas, the model denoises from the trajectories initialised by the leapfrog initialiser, which significantly decreases the required number of denoising steps and accelerates the inference process. \citet{yang2024following} built a motion memory bank by clustering trajectories in the training set as prior guidance for the diffusion model. Motion patterns and target distributions in the bank are retrieved based on the proposed addressing mechanism to guide the diffusion model for prediction. 

\subsubsection{Transformers} 
Transformers utilise attention mechanisms to learn correlations, which can capture global information and long-range dependencies effectively. For sequence data, transformers are often viewed as competitive rivals of more traditional neural network such as RNNs. This drives the application of transformers in human trajectory prediction. \citet{giuliari2021transformer} proposed new transformer networks for trajectory forecasting, challenging the dominance of LSTM models in the field. The authors explored two variants: the original transformer network~\citep{vaswani2017attention} and bidirectional transformers~\citep{devlin2018bert}, focusing on predicting individual trajectories without modelling human-human or human-scene interactions. Despite this simplicity, their transformer-based models outperformed more complex techniques, achieving state-of-the-art results on benchmarks such as Trajnet~\citep{sadeghian2018trajnet} and ETH/UCY~\citep{pellegrini2009you,lerner2007crowds}.

To simultaneously model the temporal and the social dimensions in trajectories, \citet{yuan2021agentformer} introduced a novel transformer model, AgentFormer, with an innovative agent-aware attention mechanism to preserve the personal identity. The AgentFormer enables direct interactions between the agents’ states across different time steps, enhancing long-range dependency modelling. Further, the authors presented a stochastic trajectory prediction model based on the AgentFormer following the scheme of CVAEs, fostering socially-aware and diverse trajectory generation. Specifically, the past trajectories are encoded by the AgentFormer encoders to provide conditions, while the AgentFormer decoders map the future trajectories onto the latent space and decode the latent variables to future predictions. \citet{geng2022spatio} designed a new attention mechanism by introducing a temporal attention module and a spatial attention module to capture motion features.

\citet{li2022graph} introduced a graph-based transformer for stochastic trajectory prediction. Multi-scale graphs are first established given the history trajectories and the scene segmentation maps. Subsequently, an encoder consisting of convolutional LSTMs (ConvLSTMs)~\citep{shi2015convolutional} extracts the features from these graphs to model motion patterns. Finally, a decoder uses the encoded features to predict the future trajectories. During decoding, the graph-based transformer, which models the human-to-human and the human-to-scene spatial interactions, estimates the distributions of the next positions. To ensure temporal consistency, the authors further proposed the Memory Replay module based on a memory graph, smoothening the estimates from the graph-based transformer. In addition, there are other examples where transformers are also combined with graphs such as~\citet{yu2020spatio,liu2023knowledge,liu2024attention}.

TUTR~\citep{shi2023trajectory} unifies social interactions and multimodal trajectory forecasting within a transformer encoder-decoder architecture. Unlike prior approaches~\citep{xu2022remember,xu2022socialvae} that rely on computationally expensive post-processing techniques, TUTR eliminates this step by introducing explicit global prediction. This is accomplished by generating common motion modes among pedestrians and using a mode-level transformer encoder to parse diverse motion patterns. The mode-level encoder takes as input the encoded past trajectories which are concatenated with the motion modes, to model the multimodality of the future trajectories. Then, there are neighbour embeddings which are passed through a social-level transformer decoder to capture the social interactions. Finally, the features from the transformer encoder and decoder are fed into a dual prediction network model to estimate multiple future trajectories and their corresponding probabilities. In a similar line of research, \citet{chen2022multimodal} used estimated goals as guidance to capture multimodal trajectories instead of motion modes in~\citet{shi2023trajectory} and proposed a goal-conditioned transformer for trajectory prediction.

PPT \citep{lin2024progressive} is a novel progressive pretext task learning framework based on transformers. The framework is designed to progressively enhance the model's ability through three tasks to capture the short-term dynamics and the long-term dependencies in pedestrian trajectory prediction. In Task 1, a transformer encoder is pre-trained to predict the next position for a trajectory of an arbitrary length to understand the short-term motion patterns. Task 2 aims to enhance the long-term dependency modelling of the transformer encoder by predicting diverse trajectory destinations. The pre-trained transformer encoder is then duplicated as two separate models: one dedicated to destination prediction and the other focused on intermediate position prediction in Task 3. The final prediction model involving the two separate models is trained by forecasting the entire future trajectories.

\subsection{Physics-Inspired Deep Learning Methods} 
Physics-inspired deep learning methods are a relatively new research trend in this field. The motivation of such research is to combine the advantages of both the traditional empirical non-data-driven methods and the aforementioned deep learning methods. Before statistical machine learning, there was a large body of research relying on empirical modelling. It tends to be empirical or rule-based methods derived via the first-principles approach: summarising observations into rules and deterministic systems based on certain fundamental assumptions on human motions. In such a perspective, social interactions can be modelled as forces in a particle system~\citep{helbing1995social} or an optimization problem~\citep{van2008reciprocal}, and individuals can be influenced by affective states~\citep{luo2008agent}. Later, data-driven methods were introduced, in which the model behaviour is still dominated by the assumptions on the dynamics, e.g. a linear dynamical system~\citep{he2020informative}. These methods have good explainability but normally lack prediction accuracy. In contrast, in all the previously mentioned deep learning methods, the prediction accuracy is massively improved. However, due to the black-box nature of deep neural networks, one key modelling effort is to decide the type/architecture of the network to capture as explicitly as possible certain aspects of pedestrian dynamics, \eg the environmental constraints, social interactions, and intrinsic stochasticity. To this end, since both explainability and accuracy are crucial in applications related to trajectory prediction, \eg autonomous driving, physics-inspired deep learning methods started to become popular since the year of 2022. One major modelling effort is to embed various physical systems that can best describe a group of pedestrians into deep learning for trajectory prediction. Due to the fact that this field is new, we review some recent research in detail and classify them based on the physics system they employ.

\subsubsection{Particle Systems with Deep Learning}
It is known that relatively sparse crowds can be described by particle systems in the classic work of the Social Force Model (SFM)~\citep{helbing1995social}. To make this kind of physical system fit data better, \citet{yue2022human} designed a new framework, Neural Social Physics (NSP), based on neural differential equations~\citep{chen2018neural,kidger2021neural}, combining deep learning with particle systems. In NSP, they proposed a new model NSP-SFM, which embeds the traditional SFM with learnable parameters into a deep neural network. This allows the model to possess explainability (from the physics component) and maintain outstanding prediction performance (from deep learning). NSP achieved state-of-the-art prediction accuracy on public datasets (SDD~\citep{robicquet2016learning}, ETH~\citep{pellegrini2009you}, and UCY~\citep{lerner2007crowds}), which are the most widely used datasets in computer vision in this area starting from Social-LSTM. Additionally, NSP is capable of providing human-understandable explanations for its predictions, which is significantly different from earlier pure deep learning methods.  Following NSP, other methods with similar perspectives have been proposed, such as \citep{mo2024pi} which integrates physics models into neural networks with the support of symbolic regression~\citep{schmidt2009distilling} for trajectory prediction, or \citep{yue2023human} which integrates a Bayesian approach into the social force models with deep neural networks. Since NSP is one of the first papers using physics as a prior for trajectory prediction, we provide more details below. 

NSP uses a function $q(t)$ to represent a trajectory, where $q^t=[p^t,\Dot{p}^t]^{\textbf{T}}$ consisting of the position $p^t \in \mathbb{R}^2$ and velocity $\Dot{p}^t \in \mathbb{R}^2$. Then, NSP models the dynamics of pedestrians as follows:
\begin{align}
    \frac{dq}{dt}(t) = f_{\theta, \phi}(t, q(t), \Omega(t), q^{t_h+t_f}, E) + \alpha_{\phi}(t, q^{t:t-M}),
\end{align}
where $f$ and $\alpha$ are neural network models with learnable parameters $\theta$ and $\phi$. The $\theta$ denotes explainable parameters, such as physics-related variables while the $\phi$ represents unexplainable parameters, such as neural network weights. In NSP, $f$ governs the dynamics by considering current states $q(t)$, neighbours $\Omega(t)$, destinations $q^{t_h+t_f}$, and the environment $E$. To capture stochasticity in trajectories, NSP introduces $\alpha$ which depends on the brief history $q^{t:t-M}$. Using Taylor's expansion on $q(t)$ gives:   
\begin{equation}
    q(t+\triangle t) \approx q(t) + \dot{q}(t) \triangle t =
    \begin{pmatrix}
    p(t)\\
    \dot{p}(t)
    \end{pmatrix}
    + \triangle t
    \begin{pmatrix}
    \dot{p}(t) + \alpha\\
    \ddot{p}(t)
    \end{pmatrix},
\label{eq:nsp_td}
\end{equation}
where $\triangle t$ represents the time interval, the $\alpha$ is presumed to solely affect the velocity, and $p(t)$ is assumed to possess the second-order differentiability. The generality of Eq. \ref{eq:nsp_td} enables it to incorporate any physics systems with second-order differentiability. The authors introduced the new model NSP-SFM within  NSP by integrating the social force model.

\begin{figure}[tb]
    \centering
    \includegraphics[width=0.9\linewidth]{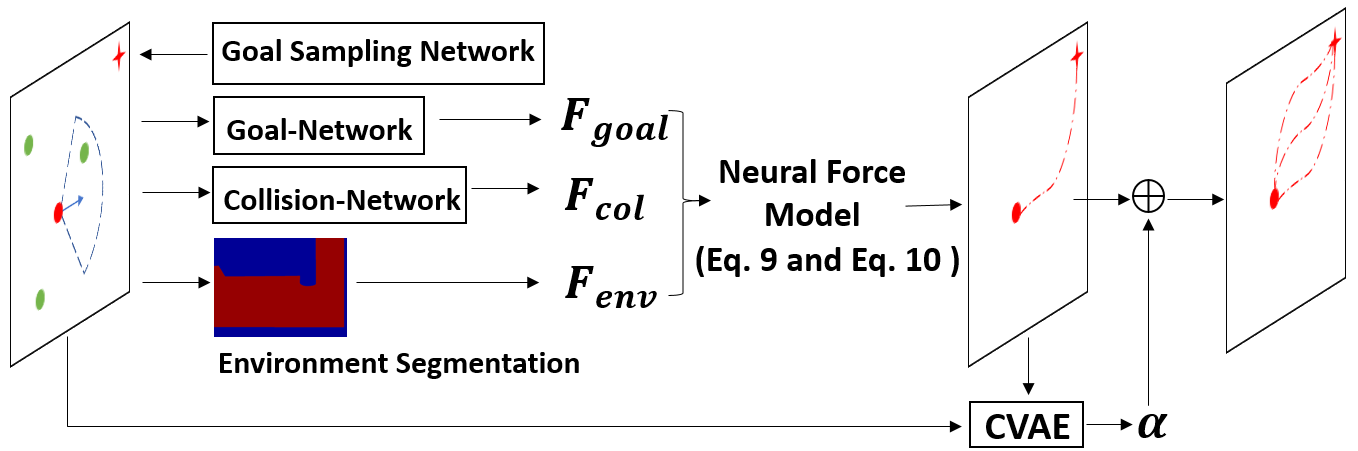}
    \caption{Overview of NSP-SFM. This figure is from~\citet{yue2022human}.}
    \label{fig:nsp-sfm}
\end{figure}
NSP-SFM focuses on modelling $\Ddot{p}(t)$ by considering three factors: goals, neighbours and the environment. As a result, we have:
\begin{align}
    \ddot{p}(t) = F_{goal} + F_{col} + F_{env},
    \label{eq:neural_force_model}
\end{align}
where $F_{goal}$, $F_{col}$, and $F_{env}$ denote the goal attraction, inter-agent repulsion and environmental repulsion, respectively. 

We show the overview of NSP-SFM in Fig.~\ref{fig:nsp-sfm}. The model estimates $F_{goal}$, $F_{col}$, and $F_{env}$ based on neural networks at each time step. The Goal Sampling Network predicts destinations used in the estimation of $F_{goal}$ given the input trajectories and the environment. Subsequently, NSP-SFM uses Eq. \ref{eq:neural_force_model} to calculate $\Ddot{p}(t)$ and adopts a semi-implicit scheme to update positions:
\begin{align}
    \dot{p}^{t+1} = \dot{p}^t + \triangle t \ddot{p}^t; \,\, p^{t+1} = p^t + \triangle t \dot{p}^{t+1}.
\end{align}
Additionally, NSP-SFM introduces stochasticity $\alpha$:
\begin{align}
    p^{t+1} = p^t + \triangle t (\dot{p}^{t+1} + \alpha^{t+1}),
\end{align}
where a CVAE is designed to predict $\alpha$. Conditioned on the historical trajectory, this CVAE takes as input the current intermediate prediction of the position and outputs $\alpha$ which is a learned random perturbation on the intermediate prediction, which gives the final prediction.

The modelling with forces inherently endows NSP-SFM with explainability. NSP-SFM retains the physical modelling in SFM and uses neural networks to predict relevant parameters which otherwise would need hand-tuning. Specifically, NSP-SFM models the goal attraction as:
\begin{align}
     F_{goal} = \frac{1}{\tau}(v_{des}^t - \dot{p}^t); \,\, \tau = NN(q^t, p^{t_h+t_f}),
\end{align}
where $v_{des}^t=\frac{p^{t_h+t_f} - p^t}{(T-t)\triangle t}$ is the desired velocity and $NN$ denotes neural networks. Given that the target agent i has the neighbour set $\Omega_i$, the estimation of the inter-agent repulsion is formulated as follows:
\begin{align}
    F_{col}^i = \sum_{j \in \Omega_i} F_{col}^{ij} =
     \sum_{j \in \Omega_i} -\nabla_{r_{ij}}\mathcal{U}_{ij}\left(\lVert r_{ij} \rVert\right) =  \sum_{j \in \Omega_i} -\nabla_{r_{ij}}r_{col} k_{ij}e^{-\lVert r_{ij}\rVert/r_{col}},
\end{align}
where $r_{ij} = p_i^t - p_j^t$, $\mathcal{U}_{ij}\left(\lVert r_{ij}\rVert\right)$ is a repulsive potential field, $r_{col}$ is a hyper-parameter, and $k_{ij} = NN(q_i^t, q^t_j)$. NSP-SFM estimates the environmental repulsion as follows:
\begin{equation}
    F_{env} = \frac{k_{env}}{\lVert p_i^t - p_{obs}\rVert} (\frac{p_i^t - p_{obs}}{\lVert p_i^t - p_{obs}\rVert}),
\end{equation}
where $k_{env}$ is a learnable parameter and $p_{obs}$ represents the positions of obstacles in the environment.

With an explicit SFM and its key parameters predicted by neural networks, NSP-SFM achieves far more accurate prediction than back then existing methods. Also, since the learn SFM is essentially a simulator, NSF-SFM can simulate more pedestrian behaviours even if the scenario is drastically different from the training data, \eg much higher densities, showing superb generalisability.

The NSP framework is flexible in the sense that the physics component is replaceable depending on what types of crowds, \eg density, are of interest. For instance, \citet{wang2024neural} directly estimated the final accelerations through a transformer structure and updated future trajectories based on Newton's laws of motion. \citet{sang2024physics} extended \citet{wang2024neural} by introducing a diffusion model to smooth predicted trajectories and a probabilistic selection module. Another instance is the particle system in NSP can be alternatively described by the time evolvement of its energy, \eg potential and kinetic, and the system tends to minimise the energy~\citep{xiang2024socialcvae}.

\subsubsection{Continuum with Deep Learning}

Although particle systems have been applied successfully in modelling relatively sparse crowds, they are not well-suited for capturing the dynamics of dense crowds. This is because the movement of dense crowds shares significantly more similarities with continuum systems such as water, compared with particle systems \citep{van2020extreme}. Therefore, it is popular to model dense crowds as continuum. Most recently, the combination of continuum dynamics and deep learning has just start to emerged in the research on dense crowds. One example is \citet{zhou2024hydrodynamics} who proposed a hydrodynamic model for crowd simulation based on the similarities between dense crowds and fluids and the unique physical and social characteristics of crowds. In their approach, the governing equation controls the crowd movement. To efficiently solve the governing equation, they designed the hydrodynamics-informed neural network for the next-step prediction. 

\begin{figure}[tb]
    \centering
    \includegraphics[width=\linewidth]{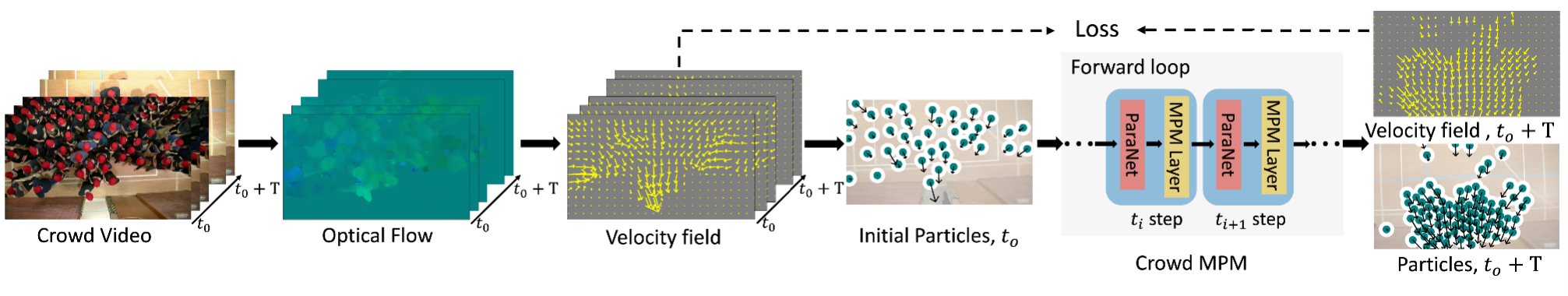}
    \caption{The overview of the method proposed in \citet{he2025learning}. This figure is from \citet{he2025learning}.}
    \label{fig:crowdmpm}
\end{figure}

Instead of seeing dense crowds as passive continumm, \citet{he2025learning} represented high-density crowds as active matter, a continuum system composed of active particles. They proposed a novel neural differential equation model called CrowdMPM to learn crowd dynamics from in-the-wild videos. The overview of the system is shown in Fig. \ref{fig:crowdmpm}. Initially, optical flows are extracted from the given video and are utilised to generate velocity fields. They can also obtain the initial individual positions (initial particles) in crowds. The proposed CrowdMPM then estimates the subsequent crowd movement. After training, the model is capable of predicting and simulating dense crowd dynamics. Furthermore, this method mitigates the scarcity of dense crowd data by utilising in-the-wild crowd videos. In these videos, tracking individuals in high-density crowds is inherently challenging. As a result, individual trajectories are not available, so that other information such as density distributions and velocity fields need to be employed. Nevertheless, the active matter modelling enables the method in \citet{he2025learning} to successfully simulate or predict fine-grained movements of dense crowds, including individual trajectories. Since CrowdMPM \cite{he2025learning} is the first method of its kind,, we would like to give more details about the method. 

CrowdMPM follows the scheme of the standard material point method (MPM) \citep{jiang2016material} while considering the features of high-density crowds, where high-density crowds are regarded as active matters. Therefore, similar to MPM, CrowdMPM is also a hybrid Eulerian-Lagrangian method, modelling dense crowd dynamics by combining strengths from the Eulerian and the Lagrangian views. Following the standard MPM, CrowdMPM begins with two governing equations   

\begin{align}
    \label{eq:governing_mass} & \frac{D\rho}{Dt} + \rho\nabla \cdot v = 0  \quad \quad \quad \quad \quad \quad \,\,\,\, \text{Conservation of mass} \\
     \label{eq:governing_mm}&\rho\frac{Dv}{Dt} = \nabla \cdot \sigma^{cm} + \rho b + f^{act} \quad \quad    \text{Conservation of momentum}
\end{align}
where $\frac{D}{Dt}$ denotes the material derivative, $\rho$ represents the density of the continuum, $v$ denotes the velocity field, $\sigma^{cm}$ is the proposed crowd material stress, $\rho b$ is the body force, and $f^{act}$ is the proposed stochastic active force. \cref{eq:governing_mass} and \cref{eq:governing_mm} denote conservation of mass and conservation of momentum, respectively. CrowdMPM is a neural differential equation model since it solves \cref{eq:governing_mass} and \cref{eq:governing_mm} by employing differential operations based on neural networks to estimate crowd dynamics.

In accordance with the standard MPM, crowdMPM solves the governing equations through a three-phase framework: 1. Particle-to-Grid transfer (P2G), 2. Grid Operations (GO), and 3. Grid-to-Particle transfer (G2P). CrowdMPM models each individual in the crowd as a particle in the Lagrangian perspective, while discretising the space using grids from the Eulerian viewpoint. In the first phase, P2G, they transfer mass and momentum from the particles to the grid to leverage the stability and computational efficiency of the Eulerian grid for solving the governing equations. Specifically, at time step n, the P2G is computed as follows:
\begin{align}
m_i^n = \sum_p w_{ip}^n m_p, \quad \quad m_i^n v_i^n = \sum_p w_{ip}^n[m_p v_p^n + m_p C_p^n(x_i-x_p^n)],
\label{eq:p2g}
\end{align}
where $m_i$ ($m_p$), $v_i$ ($v_p$), $x_i$ ($x_p$) denote the mass, velocity and position of the grid node $i$ (particle $p$), $w_{ip}$ represents weights, and $C_p$ is the affine velocity gradient. The time superscript $n$ of $m_p$ and $x_i$ are omitted because the mass of particles does not change over time, and the Eulerian grid is reconstructed at each iteration. The weights are calculated as $w_{ip}^n = \phi(x_i - x_p^n)$, where $\phi$ is a quadratic B-spline function.

Subsequently, the governing equations are solved to update the velocity on nodes in the GO phase:
\begin{align}
    \label{eq:go_1} &\hat{v}_i^{n+1} = v_i^{n} + \Delta t \frac{f_i^n}{m_i}, \\
    \label{eq:go_2} &v_i^{n+1} = BC(\hat{v}_i^{n+1}) = \hat{v}_i^{n+1} - \gamma N \left< N,\hat{v}_i^{n+1}\right>,
\end{align}
where $\Delta t$ represents the time interval, $f_i$ is the force on the node $i$, $BC$ denotes boundary conditions, $N$ denotes the surface normal vector of the boundary, and $\gamma$ is a scalar. $f_i$ is calculated as:
\begin{equation}
\label{eq:forces}
    f_i=f_i^{st} + f_i^{act} + f_i^{bd},
\end{equation}
where $f_i^{st}$, $f_i^{act}$, and $f_i^{bd}$ are the surface force (stress-based), active force and body force, respectively. Velocities updated via \cref{eq:go_1} are then corrected by imposing boundary conditions as in \cref{eq:go_2}.

Finally, G2P transfers the updated grid results back to the particles, updating their velocities and positions. Specifically, the velocities and positions of the particles for the next step are calculated using the following equations:
\begin{align}
    &v_p^{n+1} = \sum_i w_{ip}^n v_i^{n+1}, \quad x_p^{n+1} = x_p^n + \Delta t v_p^{n+1}, 
    \label{eq:g2p_1}
\end{align}
Additionally, in this phase, the affine velocity gradient $C_p$ and the deformation gradient $F_p$ are updated for the next iteration as follows:  
\begin{align}
    C_p^{n+1} = \frac{4}{\Delta x^2}\sum_i w_{ip}^{n} v_i^{n+1}(x_i- x_p^{n})^{T}, \quad F_p^{n+1} =(I + \Delta t C_p^{n+1}) F_p^{n},
    \label{eq:g2p_2}
\end{align}
where $\Delta x$ is the cell size of the grid, and $I$ is the identity matrix. Note that the deformation gradient $F_p$ is used in GO to estimate $f_i^{st}$.  

So far, what has been introduced above is a standard MPM procedure. To modify it to model high density crowds as active matters, CrowdMPM has two key technical novelties. This first technical novelty is a new stress-strain model customised for high density crowds. CrowdMPM designs new $f_i^{st}$, $f_i^{act}$, and $f_i^{bd}$ to capture the characteristics of high-density crowds, in Eq. \ref{eq:forces}. The surface force $f_i^{st}$ models three distinctive properties of dense crowds that differ from common materials: elastic asymmetry, exponential resistance and compression dominance.


Another key technical novelty of CrowdMPM is that it further incorporates active forces to represent the self-propelled nature of individuals within crowds. For this purpose, the Toner-Tu equation \citep{toner1995long} is introduced to represent the external force $f_i^{act}$ in Eq. \ref{eq:forces}. Since it contains parameters, neural networks are proposed to estimate these parameters during learning. Please refer to the paper for more details.



\subsubsection{Individual Motions in High-Density Crowds}
Very recently, there has been a new line of research, looking into predicting detailed pedestrian full-body motions in high-density crowds. Albeit still aiming at predicting trajectories, the research aims to predict detailed individual joint trajectories for the whole body for a single person, especially under unexpected physical perturbations such as pushing. Compared with all the aforementioned research, this new line of research is significantly more challenging in two folds. First, it stops treating each person as a 2D disc or particle. Instead, it models the full body for each individual. Second, it also considers potential physical interactions between pedestrians, in the presence of push and push propagation within densely packed areas. Despite presenting significant challenges, the modelling of the detailed physical interactions brings significant benefits as it allows people to predict how individuals would react in a highly dense crowd, \eg whether there is likely a fall of an individual, which could cause life-threatening danger.

\begin{figure}[tb]
    \centering
    \includegraphics[width=\linewidth]{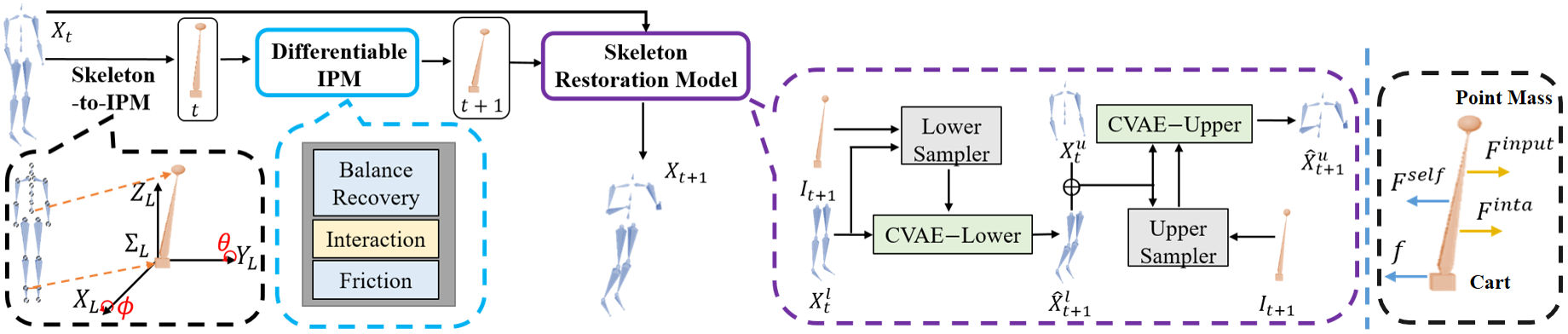}
    \caption{Overview of LDP. This figure is from~\citet{yue2024human}.}
    \label{fig:ldp}
\end{figure}

\citet{yue2024human} is probably the first work in this area to our best knowledge. \citet{yue2024human} proposed a new task: predicting 3D full-body motions under unexpected physical perturbations, to study the physical interactions and interaction propagations in crowds. More precisely, given the initial poses and input force, this task aims to predict the reactive individual motions. To this end, the authors proposed a novel latent differentiable physics (LDP) model. The overview of LDP is shown in Fig. \ref{fig:ldp}. The 3D pose $X_t$ is first simplified as an inverted pendulum model (IPM) $I_t$ by the Skeleton-to-IPM module. Then, a differentiable IPM simulates forward to obtain $I_{t+1}$ based on the 3D IPM motion equation:
\begin{align}
      M(I_t, l_{t})\Ddot{I}_t + C(I_t, \Dot{I}_t, l_{t}) + G(I_t, l_{t}) = F^{net}_t,
      \label{eq:ipm_motion}
\end{align}
where $l_t$ is the length of the rod in IPM, $M\in\mathbb{R}^{4\times4}$ represents the inertia matrix, $C\in\mathbb{R}^{4\times1}$ denotes the Centrifugal matrix, $G\in\mathbb{R}^{4\times1}$ represents external forces like gravity, and $F^{net}_t$ denotes the net fore. $M$, $C$ and $G$ can be calculated given the IPM state $I_t$, its first-order derivative $\Dot{I}_t$ and the rod length $l_t$. LDP considers the balance recovery force, interaction force, ground friction, and input force to calculate the net force $F_t^{net}$ at each step. Finally, $I_{t+1}$ is mapped onto the 3D pose $X_{t+1}$ via the Skeleton Restoration Model consisting of CVAEs and MLP samplers.        

The core idea of LDP is to model physical interactions and predict motions in the simplified IPM space. Therefore, we introduce the proposed differentiable IPM in detail. The IPM in LDP, illustrated in \cref{fig:ldp} right, consists of a cart, a massless rod and a point mass. Together, the rod and point mass form an inverted pendulum, which is mounted to the cart. At time step $t$, $I_t = [x_t,y_t,\theta_t,\phi_t] \in \mathbb{R}^4$ is used to represent the state of an IPM. $[x_t,y_t]$ denotes the coordinate of the cart moving in the XY-plane. $\theta_t$ ($\phi_t$) represents the rod's rotation angle about the $Y_L$-axis ($X_L$-axis) in the local coordinate system $\Sigma_L$. Solving \cref{eq:ipm_motion} results in $\Ddot{I}_t$ to update the IPM state through a semi-implicit framework, \ie $\Dot{I}_{t+1} = \Dot{I}_t + \triangle t \Ddot{I}_t$ and $I_{t+1} = I_t + \triangle t \Dot{I}_{t+1}$, where $\triangle t$ is the fixed time interval. Note that the rod length $l_t$ is changeable to enhance the representation ability of the IPM. LDP updates $l_t$ at each time step through an MLP:
\begin{align}
    \triangle l_t = MLP([\theta_t, \phi_t, \Dot{x}_t, \Dot{y}_t, \Dot{\theta}_t, \Dot{\phi}_t, F_t^{self}, M, l_t]), \quad l_{t+1} = l_t + \triangle l_t,
\end{align}
where $F_t^{self}$ is the balance recovery force and $M$ denotes the mass of the target person.    

The key challenge in solving the motion equation \cref{eq:ipm_motion} lies in how to estimate the net force $F_t^{net}$. LDP models the net force $F_t^{net}$ as:
\begin{align}
    F_t^{net}& = F_t^{self} + F_{t}^{inta} + f_t + F_t^{input}, 
    \label{eq:netf}
\end{align}
where $F_{t}^{inta}$, $f_t$, and $F_t^{input}$ denote the interaction force, ground friction and input force, respectively. In the data \citet{yue2024human} used, $F_t^{input}$ is given. The complexity of modelling is from $f_t$, $F_t^{self}$, and $F_{t}^{inta}$, where $f_t$ is the simplest, and $F_t^{self}$ and $F_{t}^{inta}$ are more complex. 

The friction is modelled as a damping force: $f_t = -\mu[\Dot{x}_t, \Dot{y}_t, 0, 0]$, where $\mu$ is a learnable parameter. So the friction force is learnable, as there is no good way to capture such data. 

Next, LDP models the balance recovery force as $F_t^{self} = F_t^{self-pd} + F_t^{self-nn}$, where $F_t^{self-pd}$ is the feed-forward torque based on proportional
derivative (PD) control. $F_t^{self-nn}$ is learned by neural networks which can be viewed as a torque correction. $F_t^{self-pd}$ drives the IPM to respond to external perturbation and recover balance. LDP adopts a widely used assumption that individuals subjected to perturbation typically try to recover to an upright position with zero linear velocity. So $F_t^{self-pd}$ is computed using:
\begin{align}
    F_t^{self-pd} &= K_pe_t + K_d\Dot{e}_t, & e_t & = s_d - s_t,  \notag\\
    s_d &= [0, 0, 0, 0], &  s_t & = [\Dot{x}_t, \Dot{y}_t, \theta_t, \phi_t], 
\end{align}
where $K_p$ and $K_d$ are the control hyper-parameters, $e_t$ is the state error in PD control, $s_d$ is the desired PD state, and $s_t$ is the current PD state. Although $F_t^{self-pd}$ can ensure the balance recovery, the predicted motions only using $F_t^{self-pd}$ are coarse and inaccurate. Therefore, LDP introduces $F_t^{self-nn}$ to correct $F_t^{self-pd}$ and obtain more accurate motion prediction, which is estimated by an LSTM:
\begin{equation}
    F_t^{self-nn} = LSTM([\theta_t, \phi_t, \Dot{x}_t, \Dot{y}_t, \Dot{\theta}_t, \Dot{\phi}_t, M]).
\end{equation}

LDP models the interaction force for each pair of individuals and uses the summation of interaction forces $F_{t,nj}^{inta}$ from all the neighbours in the neighbourhood $\Omega_{t,n}$ of an individual as the final interaction $F_{t,n}^{inta}/F_{t}^{inta}$ for the n$th$ person: 
\begin{equation}
F_{t,n}^{inta} = \sum_{j \in \Omega_{t,n}} F_{t,nj}^{inta} = \sum_{j \in \Omega_{t,n}} F_{t,nj}^{inta-bs} + F_{t,nj}^{inta-nn},
\label{eq:inta}
\end{equation}
where $\Omega_{t,n}$ denotes the set of neighbours of the n$th$ person at time $t$. As \cref{eq:inta} shows, each $F_{t,nj}^{inta}$ consists of the basic interaction force $F_{t,nj}^{inta-bs}$ and the neural interaction force $F_{t,nj}^{inta-nn}$. $F_{t,nj}^{inta-bs}$ models the primary interaction, while $F_{t,nj}^{inta-nn}$ serves as a supplementary component. To model $F_{t,nj}^{inta-bs}$, LDP adopts different strategies to capture interactions in $[x,y]$ and $[\theta, \phi]$, resulting in forces $F_{nj}^{bs-xy}\in\mathbb{R}^2$ and $F_{nj}^{bs-\theta \phi}\in\mathbb{R}^2$, so that $F_{t,nj}^{inta-bs} = [F_{nj}^{bs-xy}, F_{nj}^{bs-\theta \phi}]$, where superscript $inta$ and subscript $t$ are omitted for simplicity. Further, LDP models $F_{nj}^{bs-xy}$ by using a repulsive potential energy function $\mathcal{U}$:         
\begin{align}
    & F_{nj}^{bs-xy}(r_{nj}) = - \nabla_{r_{nj}} \mathcal{U}[b(r_{nj})],
\end{align}
where $r_{nj} = r_n - r_j$ is the relative position of carts within the XY-plane between the n$th$ target person and their neighbour j. $\mathcal{U}[b] = ue^{-\frac{b} {\sigma}}$ has elliptical equipotential lines, where $u$ and $\sigma$ are hyper-parameters. $b$ represents the semi-minor axis of the ellipse and is calculated as:
\begin{align}
   b = \frac{1}{2}\sqrt{(\lVert r_{nj} \rVert +  \lVert r_{nj} - \triangle t \Dot{r}_{jn} \rVert)^2 - \lVert \triangle t \Dot{r}_{jn} \rVert ^ 2 },
\end{align}
where $\Dot{r}_{jn} = \Dot{r}_j - \Dot{r}_n$.

LDP adopts a different strategy to estimate $F_{nj}^{bs-\theta \phi}$. Overall, they are estimated based on the relative orientation of a pair of IPMs. Finally, the neural interaction force $F_{t,nj}^{inta-nn}$ is learned by an MLP:       
\begin{equation}
    F_{t,nj}^{inta-nn} = MLP([x_{nj}, y_{nj}, \theta_n, \phi_n, \theta_j, \phi_j, \Dot{x}_{nj}, \Dot{y}_{nj}, \Dot{\theta}_{nj}, \Dot{\phi}_{nj}]),
\end{equation}
where the definition of $x_{nj}$ and $y_{nj}$ ($\Dot{x}_{nj}$, $\Dot{y}_{nj}$, $\Dot{\theta}_{nj}$ and $\Dot{\phi}_{nj}$) follows an approach similar to $r_{nj}$ ($\Dot{r}_{jn}$).

\subsection{Conclusion}
\begin{table}[tb]
    \centering
    \caption{The comparison of key techniques in human trajectory prediction.}
    \begin{tabular}{c|c|c|c|c}
    \hline      
    Technique  & \, Accuracy  \, &  \, Explainability  \, &  \, Data Requirement  \, &  \, Computational Cost  \,    \\
    \hline
    Traditional  & Low-Medium & High & Low & Low \\
    \hline
    RNN  & High & Low & High & Medium \\
    \hline
    CNN  & High & Low & High & Medium \\
    \hline 
    GNN & High & Medium & High & High \\
    \hline 
    Generative Model  & Very High & Low & High & High \\
    \hline
    Transformer & Very High & Low & High & High \\
    \hline
    Physics-inspired & \multirow{2}{*}{Very High} & \multirow{2}{*}{High} & \multirow{2}{*}{Medium} & \multirow{2}{*}{Medium}\\
    Deep Learning & & & & \\
    \hline 
    \end{tabular}
    \label{tab:cmp_pred}
\end{table}

The introduction of deep learning has significantly advanced the research in human trajectory prediction. This section first briefly reviewed early traditional machine learning methods and then systematically covered the mainstream deep learning approaches. To give a high-level comparison of their advantages and drawbacks, we examine them by several indicators, Accuracy, Explainability, Data Requirement, and Computational Cost, shown in Table \ref{tab:cmp_pred}. Note this is a qualitative comparison as the evaluation metrics and experimental settings vary greatly across different publications. Direct quantitative comparison between them across all indicators is itself an interesting, yet open research question.

Traditional machine learning methods typically achieve low to medium prediction accuracy but offer strong explainability. In contrast, deep learning approaches generally excel in prediction accuracy. Although most deep learning methods struggle with explainability, physics-inspired deep learning methods mitigate this limitation and provide substantial explainability. The performance of deep learning models often depends heavily on the volume of available data, resulting in higher data requirements. However, the incorporation of physical priors effectively reduces the data requirement for physics-inspired deep learning methods. Moreover, due to their large number of trainable parameters, deep learning methods typically need more computational resources compared to traditional machine learning methods, particularly when it comes to the methods based on GNNs, generative models or transformers. In conclusion, physics-inspired deep learning methods demonstrate the strongest overall performance, particularly in accuracy and explainability, motivating further exploration of the type of methodologies.

\begin{table}[t]
    \centering
    \caption{The comparison of different methods on prediction accuracy. The best results are in bold, and the second-best results are underlined. Methods with * focus on deterministic trajectory prediction. The others are designed for stochastic trajectory prediction, where 20 future trajectories are generated, and the minimal error is reported. N/A stands for not applicable. }
    \begin{tabular}{c|c|cc|cc}
    \hline      
    \multirow{2}{*}{Technique} &  \multirow{2}{*}{Method}  & \multicolumn{2}{c|}{ETH/UCY} & \multicolumn{2}{c}{SDD} \\
    \cline{3-6}
    & & \multicolumn{1}{c|}{\,ADE\,} & \,FDE\, & \multicolumn{1}{c|}{\,ADE\,} &\, FDE\, \\
    \hline
    Traditional & \, Physics-based Model*  \citep{yamaguchi2011you} \, & \multicolumn{1}{c|}{0.55}  & 0.99 & \multicolumn{1}{c|}{36.48} & 58.14\\
    \hline
    
    \multirow{2}{*}{RNN} & \, Social-LSTM*  \citep{alahi2016social} \,  & \multicolumn{1}{c|}{0.72} & 1.54 & \multicolumn{1}{c|}{31.19} & 56.97\\
    \cline{2-6}
    & \, Group-LSTM*  \citep{zhang2019sr} \, & \multicolumn{1}{c|}{0.45} & 0.94 & \multicolumn{1}{c|}{N/A} & N/A \\
    \hline
    \multirow{2}{*}{CNN} & \, TrajCNN*  \citep{nikhil2018convolutional} \, & \multicolumn{1}{c|}{0.59} & 1.22 & \multicolumn{1}{c|}{N/A} & N/A \\
    \cline{2-6}
    & \, Ped-CNN*  \citep{zamboni2022pedestrian} \, & \multicolumn{1}{c|}{0.44} & 0.91 & \multicolumn{1}{c|}{N/A} & N/A \\
    \hline
    \multirow{2}{*}{GNN} &  \, Social-STGCNN  \citep{mohamed2020social} \,  & \multicolumn{1}{c|}{0.44} & 0.75 & \multicolumn{1}{c|}{20.6} & 33.1 \\
    \cline{2-6}
    & \, D-STGCN  \citep{sighencea2023d} \,  &\multicolumn{1}{c|}{0.42} & 0.68 &\multicolumn{1}{c|}{15.18} & 25.50 \\
    \hline
    \multirow{3}{*}{Generative Model} & \, Social-GAN  \citep{gupta2018social} \, & \multicolumn{1}{c|}{0.58} &  1.18 & \multicolumn{1}{c|}{27.23} & 41.44\\
    \cline{2-6}
        & \, MLD  \citep{wu2024motion} \, & \multicolumn{1}{c|}{\underline{0.17}} & 0.35 & \multicolumn{1}{c|}{6.61}  & 12.65 \\
    \cline{2-6}

    & \, IDM \citep{liu2025intention} \, & \multicolumn{1}{c|}{0.18} & \underline{0.27} & \multicolumn{1}{c|}{\textbf{6.38}} & 11.02 \\
    \hline
    \multirow{2}{*}{Transformer} & \, AgentFormer  \citep{yuan2021agentformer} \,   & \multicolumn{1}{c|}{0.18} & 0.29  & \multicolumn{1}{c|}{N/A} & N/A\\
    \cline{2-6}
    & \, PPT \citep{lin2024progressive} \, & \multicolumn{1}{c|}{0.20} & 0.31 & \multicolumn{1}{c|}{7.03} & \underline{10.65}\\

    \hline
    Physics-inspired & \, NSP-SFM  \cite{yue2022human} \, & \multicolumn{1}{c|}{\underline{0.17}} & \textbf{0.24} & \multicolumn{1}{c|}{\underline{6.52}} & \textbf{10.61}\\
    \cline{2-6}
    Deep Learning & NDCPM \citep{wang2024neural} &\multicolumn{1}{c|}{\textbf{0.15}} & 0.33 & \multicolumn{1}{c|}{N/A} & N/A \\
    \hline
    \end{tabular}
    \label{tab:acc_pred}
\end{table}

Although it is difficult to directly compare the reviewed methods in the aforementioned four indicators, it is still possible to compare them in their predictive capabilities due to most of the reviewed deep learning methods share common datesets and evaluation metrics for trajectory prediction.

A detailed numerical comparison of prediction accuracy is shown in Table \ref{tab:acc_pred} based on the original results reported in the publications. Specifically, we highlight the performance of representative methods which are either pioneering or demonstrating excellent prediction accuracy within each category. This comparison is based on evaluations conducted on two widely used public datasets: ETH/UCY \citep{pellegrini2009you,lerner2007crowds} and SDD \citep{robicquet2016learning}. The primary evaluation metrics used are Average Displacement Error (ADE), which measures the average discrepancy between the predicted trajectory and its ground truth, and Final Displacement Error (FDE), which quantifies the error at the final predicted position. The specific experimental setting is to predict the future trajectory $\{p^t\}_{t=t_p+1:t_p+t_f}$ for the next $t_f$ timesteps given the past trajectory $\{p^t\}_{t=1:t_p}$, where $p^t$ denotes the 2D position and each timestep has 0.4 seconds. ADE and FDE are calculated as:
\begin{align}
    ADE = \frac{1}{Nt_f}\sum_{i=1}^N\sum_{t=t_p+1}^{t_p+t_f} \left \| p_i^t - \hat{p}_i^t \right \|_2, \quad FDE = \frac{1}{N}\sum_{i=1}^N \left \| p_i^{t_p+t_f} - \hat{p}_i^{t_p+t_f} \right \|_2,
\end{align}
where $p_i^t$ is the 2D position for the $i$th trajectory, $\hat{p}_i^t$ is the predicted position, and $N$ is the total amount of trajectories. Following standard practice in the literature \citep{gu2022stochastic,yue2022human,shi2023trajectory}, $t_p=8$ and $t_f=12$. The ADE and FDE results are reported in meters for ETH/UCY and pixels for SDD. Notably, employing deep learning techniques such as RNNs, CNNs, and GNNs has substantially improved prediction accuracy, reducing ADE and FDE by approximately 20\% and 30\%, respectively, on ETH/UCY, and by about 50\% for both metrics on SDD, compared with traditional machine learning approaches. The adoption of generative models and transformer architectures has further significantly enhanced the prediction performance across all metrics and datasets. Ultimately, physics-inspired deep learning methods achieve the highest accuracy. In summary, generative models, transformers, and particularly physics-inspired deep learning methods demonstrate superior prediction accuracy for human trajectory forecasting.

\section{Crowd Behaviour Recognition}

Crowd behaviour recognition is the process of analysing and identifying patterns or activities in the movements or behaviours of crowds. It aims to understand how individuals within a crowd interact and respond to their environments, other individuals, and external stimuli. This chapter reviews crowd behaviour recognition research based on machine learning from crowd videos. Similar to trajectory prediction, traditional machine learning played a key role in behaviour recognition, then deep learning has been becoming popular recently. 

\subsection{Traditional Machine Learning Methods}
Traditional machine learning methods typically involve manually designing effective features to represent crowd dynamics, followed by training a classifier to recognise crowd behaviours based on these features. These traditional methods generally model crowd movements from two perspectives: holistic and individual-based.

Holistic methods treat the crowd as a single entity. Often this is because the individual movements either cannot be accurately detected \eg due to the view blocked in the camera, the camera being too far away from the crowds, or are regarded as insignificant when it comes to the identification of the global behaviour pattern. Therefore, these approaches focus on the macroscopic behavioural patterns and often treat the crowds as a continuum, which are well-suited for analysing dense crowds where each person occupies merely a small number of pixels hence difficult to detect/track. \citet{solmaz2012identifying} proposed a framework for identifying crowd behaviours using stability analysis of dynamical systems. Their approach employs a Lagrangian perspective and overlays a grid of particles onto a scene. This way, they defined a dynamical system based on the particle movements, which is indicated by the optical flows of the video. The dynamical system is not fully parameterised but approximated by a Taylor's expansion of the system with the first-order term where the Jacobian matrix dictates the dynamics. By analysing the particle trajectories and employing the eigenvalues of the Jacobian matrix, the method classifies specific behaviours (Lane, Blocking, Bottleneck, Fountainhead, and Arch/Ring) through stability patterns. In comparison, \citet{su2013large} took an Eulerian perspective and proposed a new spatiotemporal viscous fluid field to model crowd dynamics for large-scale crowd behaviour recognition. This approach constructs the novel spatiotemporal features, by incorporating both the appearance variations of the crowds and the interaction forces among pedestrians estimated through the shear stress in the fluid field. Then it employs a latent Dirichlet allocation model~\citep{blei2003latent} to identify crowd behaviours such as dispersion and gathering based on these spatiotemporal features. More recently, \citet{matkovic2022new} presented a novel quantum mechanics-inspired method to recognise dominant motion patterns in macroscopic crowd analysis. The method extracts optical flows from the input video and introduces particles similar to~\citet{solmaz2012identifying}. Then, it constructs a wave field based on the optical flows, the particles, and some pre-defined wave functions. The method defines the peak of the wave field as a meta-tracklet, which indicates the most probable particle flow. Eventually, a classifier is employed to recognise the dominant motion patterns such as Inline and Circle, based on the meta-tracklets and the functions of fuzzy predicates.  

Individual-based methods regard the crowd as a collection of individuals, enabling detailed analysis of behaviours at the microscopic level. \citet{zhou2012understanding} used agents with a linear dynamic system to model the motion of each pedestrian. They proposed a novel mixture model of dynamic agents to analyse the collective crowd behaviours. After learning from the data, the mixture model can simulate and classify crowd behaviours. \citet{choi2012unified} introduced a framework which unifies and integrates multi-target tracking and crowd behaviour recognition. They proposed a novel hierarchical graphical model to jointly optimise the target tracking and the activity classification. The model links three levels of activities (\ie individual, interaction, and crowd) and leverages the contextual information at each level to improve recognition. The clustering-based methods such as \citet{wang2009unsupervised,zhou2012understanding} later inspired more fine-grained behavioural analysis on crowd activities. \citet{wang2016path,wang2017trending} extended \citep{wang2009unsupervised} into more detailed spatial pattern recognition for crowd activities. \citet{wang2016globally} extended \citet{wang2017trending} and proposed new unsupervised spatio-temporal behaviour recognition. Finally, \citet{he2020informative} extended \citep{wang2016globally} into unsupervised behaviour analysis based on space, time and dynamics simultaneously.

\subsection{Deep Learning Methods}
Similar to human trajectory prediction, deep learning has made a big impact on the research in crowd behaviour recognition, by overcoming the limitations of traditional approaches, enabling the analysis of complex scenarios with greater accuracy, scalability, and robustness. Deep learning methods for crowd behaviour recognition utilise neural networks to implicitly learn effective features from data, greatly minimising the dependence on manual feature engineering. Additionally, these methods tend to integrate feature extractors with classifiers, providing efficient end-to-end models. With the advancement of deep learning, CNNs, RNNs, GNNs, transformers \etc. have been explored to identify various crowd behaviours. Similar to the above section, we also classify these methods based on the types of neural networks.

\subsubsection{Convolutional Neural Networks}

CNNs are widely used in crowd behaviour recognition because they are well-suited for processing image and video data, which is the primary source of information in this domain (\eg CCTV or surveillance videos). \citet{shao2015deeply} proposed a novel deep model consisting of two CNN branches, with the same architecture, which integrates the appearance and the motion features to jointly learn discerning features for crowd behaviour recognition. One branch of the model takes a frame of the video to capture the static features of the crowd, while the other branch accepts the continuous motion maps derived from the crowd video to capture the dynamic features. Furthermore, inspired by the behavioural features (stability, conflict, and collectiveness) widely employed in prior studies, the authors proposed three continuous motion maps representing various dynamics information. Then the features from the two branches are fused to serve as the input of a classifier, which is a fully connected layer. Extensive experiments demonstrated that the proposed model had superior performance over handcrafted features and state-of-the-art baselines. Moreover, the authors built a large-scale dataset, WWW Crowd Dataset, containing 10,000 crowd videos from 8,257 scenes. As data is the foundation of deep learning, the data contribution is considered a distinctive contribution to the field. \citet{shao2015deeply} is the pioneering work applying deep learning in crowd behaviour recognition, with a new large-scale dataset, which has inspired a stream of research~\citep{shao2016slicing,ullah2019two,deng2020behavior}. Therefore, we provide more details of the method. 

\begin{figure}[tb]
    \centering
    \includegraphics[width=0.95\linewidth]{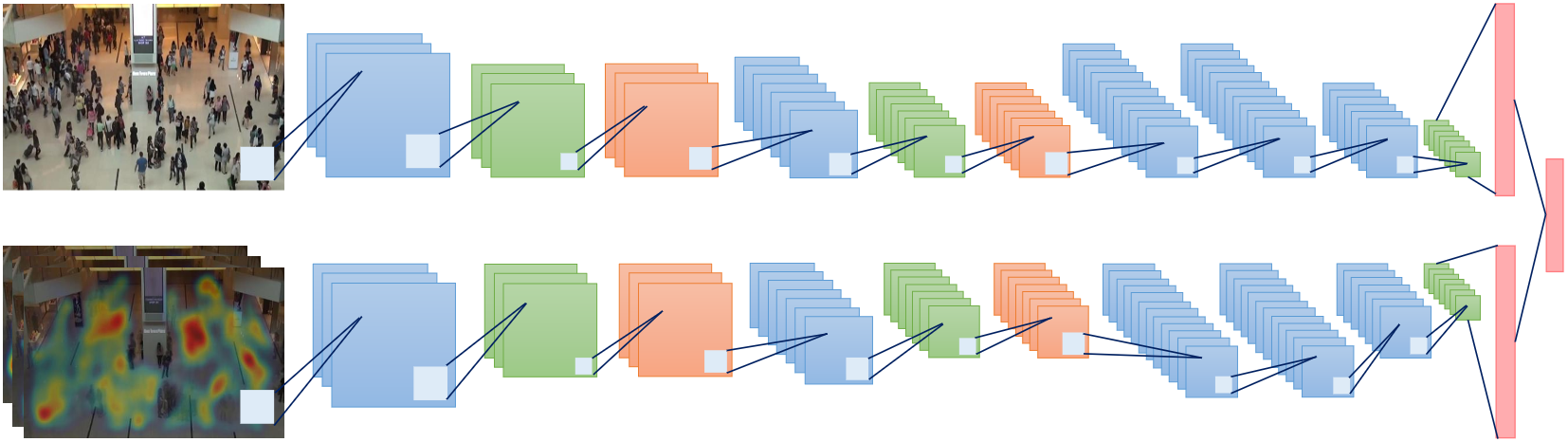}
    \caption{The overview of the proposed deep model in \citet{shao2015deeply}. Blue, green, orange, and red blocks denote convolutional, pooling, normalization, and fully connected layers, respectively. This figure is from \citet{shao2015deeply}.}
    \label{fig:shao}
\end{figure}
The overview of the proposed model in \citet{shao2015deeply} is shown in \cref{fig:shao}. A single frame and continuous motion maps are extracted from the given video to serve as input on the left. Each branch has a CNN constructed by stacking convolutional, pooling, normalization, and fully connected layers, with a rectified linear unit applied after every convolutional layer as the activation function. The final fully connected layer is the classifier followed by a sigmoid activation function, which predicts the probability of the input data showing a certain behaviour. The calculation of three continuous motion maps is inspired by~\citet{zhou2013measuring,shao2014scene}. Specifically, the authors first detected the tracklets of motions and established a 10-nearest-neighbour graph for all tracklets. Then they used a stability descriptor by averaging over the unchanging neighbours for each node within the graph. Next, they proposed a conflict descriptor based on the graph's velocity correlation of nearby nodes. As for collectiveness, the descriptor from~\citet{zhou2013measuring} was employed. Finally, they could generate a stability/conflict/collectiveness descriptor map for each frame. Averaging the descriptor maps along the temporal dimension produced three motion maps. To enrich the representation of motion information, these motion maps were interpolated to generate continuous motion maps.

\citet{ullah2019two} explored crowd behaviour recognition following the modelling idea in \citet{shao2015deeply}, providing a two-stream model based on CNNs. Both methods use the frames of the crowd video to capture the appearance information, while \citet{ullah2019two} utilised optical flows to capture motion information. Finally, softmax scores from the two streams are fused to obtain classification results. The classic two-stream architecture is also used to explore crowd behaviour recognition in other CNN-based methods~\citep{shuaibu2017adaptive,wei2020very,ullah2021multi}.

\citet{mandal2018deep} proposed a novel crowd behaviour recognition model that combines deep residual neural networks with subclass discriminant analysis~\citep{zhu2006subclass} to enhance feature extraction and classification accuracy. The model uses a ResNet-50~\citep{he2016deep} to extract features from each frame of the input video. To obtain the intra-class variances, the authors introduced spatial partition trees~\citep{wang2013trinary} which segment each crowd behaviour class into subclasses within the feature maps. Subsequently, the features from these subclasses are further refined using eigenvalue-based regularisation to extract more discriminative features. Next, the model computes a total subclass scatter matrix from the refined features and applies the matrix to extract the final low-dimensional features. Finally, a 1-nearest-neighbour classifier is employed to recognise crowd behaviours based on the low-dimensional features.  

PIDLNet~\citep{behera2021pidlnet} combines CNNs with physics-related metrics to characterise crowd videos. Specifically, the model introduces the inflated 3D ConvNets (I3D)~\citep{carreira2017quo}, a video classification network, to produce implicit features by capturing and aggregating the appearance and the motion features from the input video. Meanwhile, PIDLNet uses the optical flow information extracted by I3D to estimate physics-based features consisting of entropy and order parameters. Then, the implicit features and the physics-based features are fused. A classifier, a linear layer, takes the fused features as input to recognise structured and unstructured crowds. 

\citet{bendali2021ensemble} integrated existing CNN models by ensemble learning to improve recognition performance. This ensemble includes I3D, two-stream I3D~\citep{carreira2017quo}, Convolutional 3D~\citep{tran2015learning}, and Resnet 3D~\citep{hara2017learning}.

3D-AIM \citep{choi2024three} is a new CNN model based on atrous convolutions~\citep{chen2017deeplab} for crowd behaviour recognition. The model integrates two key components: an atrous block with atrous convolutions which captures the spatial features, and an inception block with standard convolutions which extracts the spatiotemporal features. 3D-AIM utilises the atrous convolutions to expand the receptive field without increasing the computational parameters when extracting the spatial features. Furthermore, the authors proposed a novel separation loss to assign greater weighting onto the difficult-to-classify examples and enhance class separation.

\subsubsection{Recurrent Neural Networks}

The importance of temporal information in crowd data motivates the usage of RNNs. RNNs enable the modelling of the dynamics of individual information and how they influence each other when interactions happen. The combination of CNNs with RNNs has further advanced crowd behaviour recognition. \citet{wang2017recurrent} proposed a hierarchical recurrent modelling approach. Specifically, a person-level LSTM is first used to capture the individual dynamics, which takes the appearance and the motion information of each individual as input at each time step. The method extracts individual information from videos using traditional methods~\citep{choi2009they,choi2012unified} and CNNs. Each person is described by the features based on the outputs of the person-level LSTM. Subsequently, a group-level LSTM is employed to model the interactions at the group level, where groups are identified according to the spatiotemporal proximity of individuals. Individual features within a group are ordered into a sequence by their spatial coordinates in the frame space and fed into the group-level LSTM. Similarly, each group is described by the group features out of the group-level LSTM. Finally, a scene-level LSTM is employed to capture the whole crowd behaviour.

\citet{vahora2019deep} proposed a deep neural network model to recognise crowd behaviours by leveraging contextual information. At each time step, a frame of the input video is fed into a CNN model which extracts scene-level features and outputs activity scores. Meanwhile, the model uses a CNN model to extract the individual features from detected individuals in the frame and employs a pooling operation to aggregate these features. An RNN model (LSTM/GRU) then receives the aggregated features as input and estimates the activity scores computed by softmax. In the end, a probabilistic inference model produces the final activity label based on all the previous softmax scores.    

\citet{yan2019crowd} presented an encoder-decoder model to recognise behaviours in crowd videos and generate corresponding captions. The CNN-based encoder extracts features from input videos. These features are fed into the RNN-based decoder to predict captions.   

\citet{behera2021characterization} explored the combination of traditional handcrafted features and RNNs. To be specific, the proposed model extracts features incorporating 2D motion histograms, order parameters, and entropy from each frame of the input video. These temporally related features are fed into an LSTM to produce classification results at every time step. Finally, all classification results are aggregated to classify the crowd as structured or unstructured.

\citet{rezaei2021real} proposed the Conv-LSTM-AE model to capture high-level representations of data, where AE stands for the autoencoder~\citep{rumelhart1986parallel}. Conv-LSTM-AE can be seen as a multi-task learning framework. Its main architecture is an autoencoder. The encoder accepts a sequence of optical flow images obtained from the input video based on~\citep{facciolo2013g} as input to extract features. The encoder consists of convolution layers and convolutional
LSTMs (ConvLSTMs)~\citep{shi2015convolutional}. Subsequently, the encoded features are fed into two separate branches. One branch is a classification branch which employs an MLP to recognise crowd behaviours. The other branch reconstructs the optical flow images, integrating ConvLSTMs and transposed convolution layers~\citep{zeiler2011adaptive}. The proposed model employs a loss function that accounts for reconstruction and classification errors during training. Prior to~\citet{rezaei2021real}, ConvLSTMs were also used to model spatio-temporal information for behaviour recognition~\citep{li2018deep}. Later, \citet{chaturvedi2024fight} combined ConvLSTMs with the attention mechanism to identify crowd activities.

\subsubsection{Graph Neural Networks}
Similar to trajectory prediction where moving pedestrians can be modelled by a graph, GNNs have unique advantages when used for crowd behavioural analysis. Their abilities to model non-Euclidean data allow individuals and objects in the environment to be seen as graph nodes and the relationship between any two nodes to be modelled by edges. Therefore, GNNs are highly effective at modelling the complex interactions and relationships inherent in crowd dynamics, facilitating the learning of both local and global patterns within a crowd. Moreover, GNNs can handle multimodal data by embedding appearance, motion, and semantic features into the graph structures, allowing a multi-faceted modelling of crowd behaviours. Therefore, a series of GNN-based methods have been proposed in this area. 

\citet{behera2021crowd} regarded crowd behaviour recognition as a graph classification problem and used a graph convolutional neural network to identify types of graphs which correspond to different behaviours. Given the crowd video, constructing graphs is to introduce a node for each group in the crowd and establish edges based on the orientations of groups. Specifically, the proposed method uses a Langevin equation-based framework~\citep{behera2021understanding} to detect groups. Then, seeing crowds as fluids, the method extracts features for each group from the video, including optical flow features and physics-related features. Combining these features gives the node features. The edges of the graph are then built if the orientation similarity between the two groups is higher than a certain pre-defined threshold. Finally, a graph convolutional neural network classifies these graphs.

\citet{liu2021multimodal} introduced a multimodal semantic context-aware graph neural network, designed to capture the visual and the semantic interactions in complex scenes. The proposed model constructs two multimodal visual sub-graphs: an RGB graph to capture the appearance cues and an optical-flow graph for motion patterns. Both sub-graphs build nodes for individuals and are fully connected. Node features are extracted from RGB and optical flow images using I3D~\citep{carreira2017quo}. Subsequently, the model designs the modality-specific and cross-modal aggregation layers to refine graph representations of two sub-graphs. Additionally, a semantic graph is built with nodes corresponding to individual actions and crowd behaviour labels, representing edges by a fixed adjacency matrix. A bi-directional mapping mechanism based on graph convolutional networks is employed to link the multimodal visual graphs to the semantic graph, enriching the visual representations with the semantic contexts. Finally, the multimodal semantic context-aware features are used to recognise the crowd activities. As a follow-up paper, \citet{liu2022visual} extended their work~\citep{liu2021multimodal} by utilizing estimated pose information of individuals in crowds to guide modality-specific and cross-modal aggregation of two sub-graphs.

\citet{longobardi2024graphic} aimed to classify diverse crowd behaviours within the same environment, unlike previous methods~\citep{shao2014scene,su2016crowd} which rely on pre-segmented scenes with single crowd behaviour. To this end, the authors proposed a novel two-stream graph neural network model. The first stream feeds a set of graphs to a graph convolutional network separately to output a set of graph representations. They used graphs of two layers, the bottom layer and the top layer, based on a grid/tile discretization of the scene. Bottom-level and top-level graphs are constructed based on detected groups and tiles, respectively. To enhance the classification, the model incorporates motion flow images, capturing both the individual and the collective dynamics. The second stream uses a CNN to extract the motion features from the motion flow images. The model then combines the graph representation with the motion features to conduct node-level classification.

\subsubsection{Transformers}


Given a sequence of input data, transformers use attention mechanisms to learn the correlation between different parts of the data and capture long-range dependencies. In crowd behavioural recognition, this provides an alternative way for a model to capture the spatial and temporal relationships in crowds. Therefore, it is no surprise that transformers have been actively employed in this area.

\citet{tamura2022hunting} proposed a new transformer-based method to recognise crowd activities. The method feeds the input video into I3D to extract a batch of multi-scale feature maps. Each feature map is then passed through an average pooling layer followed by some projection convolution layers to reduce the computational demands of transformers, resulting in altered feature maps. Subsequently, a deformable transformer encoder~\citep{zhudeformable} receives the altered feature maps and a collection of multi-scale position encodings~\citep{zhudeformable} as input to produce refined feature maps. Finally, a set of learnable query embeddings and the refined feature maps are fed into a deformable transformer decoder~\citep{zhudeformable} to obtain effective behaviour representations. \citet{tamura2024design} extended this work by employing two separate deformable transformer decoders, one for individuals and another for groups. Each decoder is equipped with its own learnable query embeddings.

\citet{zuo2023v3trans} proposed another transformer to capture effective features for crowd behaviour recognition. The model processes the input video into a sequence of tokens and maps them onto embedding vectors by a linear layer, following the framework of Visual Transformers~\citep{han2020survey}. Then, a transformer accepts these embedding vectors as input to extract features. The multi-head self-attention module in the transformer is modified to contain both spatial and temporal blocks. Finally, the extracted features are fed into an MLP classifier to identify the crowd behaviours.      

\begin{figure}[tb]
    \centering
    \includegraphics[width=\linewidth]{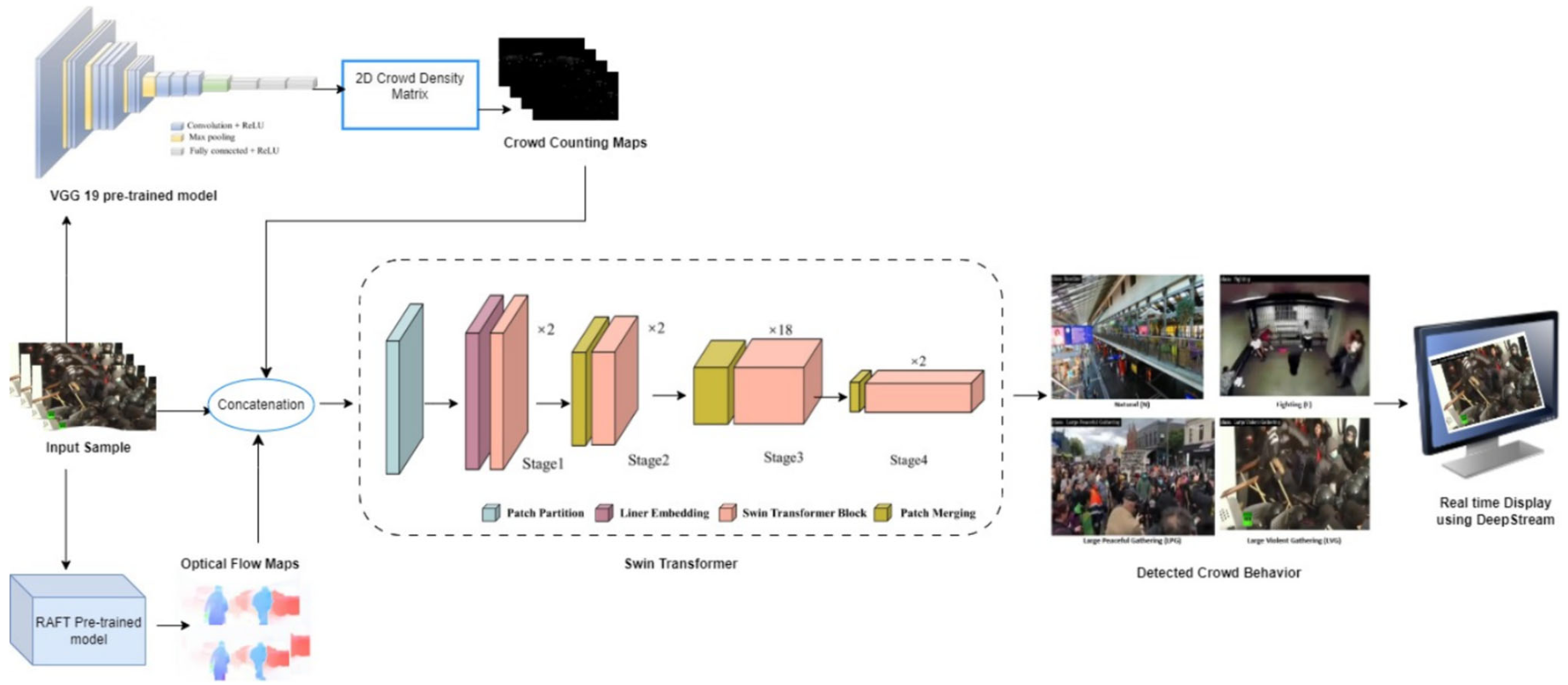}
    \caption{Overall structure of the proposed model in~\citet{qaraqe2024crowd}. This figure is from~\citet{qaraqe2024crowd}.}
    \label{fig:qaraqe}
\end{figure}

\citet{qaraqe2024crowd} provided an in-depth exploration of crowd behaviours varying in size and levels of violence. The authors categorised crowd behaviours into natural, large peaceful gathering, large violent gathering, and fighting. They proposed a swin transformer-based model to classify behaviours. The overall structure of the proposed model is shown in \cref{fig:qaraqe}. Specifically, crowd counting and optical flow maps are extracted from a given video. Then, these features are concatenated with the given video to form the input for a swin transformer proposed by \citet{liu2022video}, which is efficient for video recognition. Finally, the transformer produces effective classification features. Comprehensive experiments demonstrated that the proposed method outperformed existing state-of-the-art approaches. Moreover, the authors created a large dataset containing 68 hours of videos from both closed-circuit television and social media. Identifying violent behaviours is crucial for crowd management and public security~\citep{saxena2008crowd}. The proposed model in~\citet{qaraqe2024crowd} has advanced the research on violence detection in crowds by applying cutting-edge deep learning techniques. Therefore, we would like to provide more details of the model.         

\begin{figure}[tb]
    \centering
    \includegraphics[width=\linewidth]{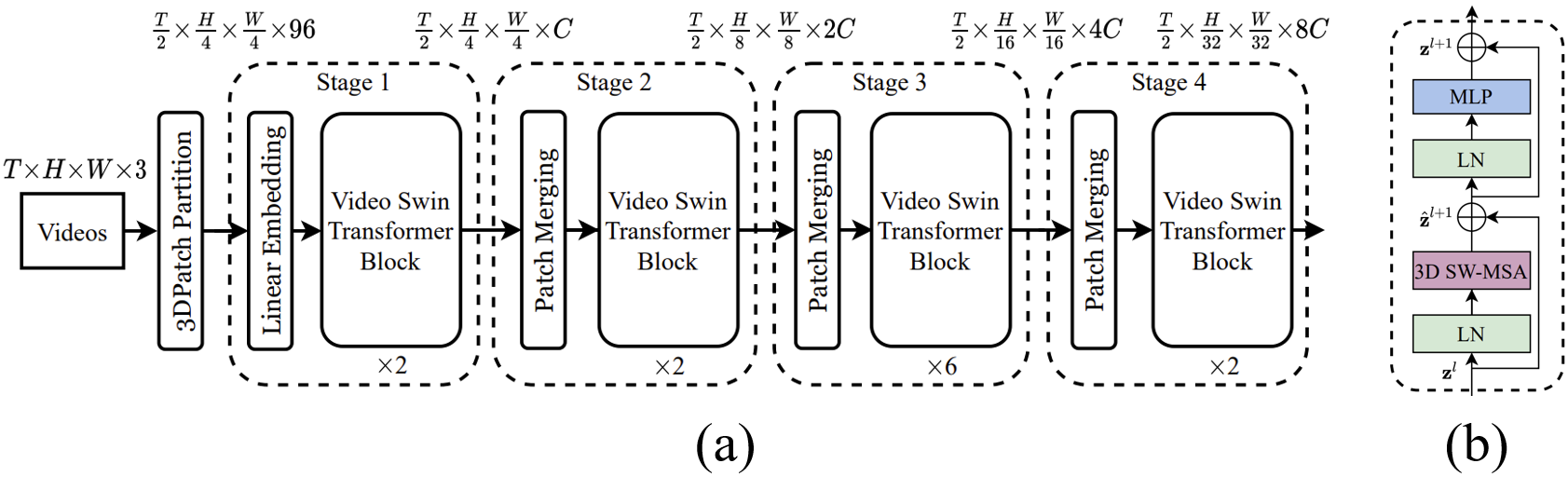}
    \caption{Architecture of the swin transformer (a) and an illustration of the video swin transformer block (b). These two figures are from \citet{liu2022video}.}
    \label{fig:swin}
\end{figure}

\citet{qaraqe2024crowd} first followed~\citet{wan2021generalized} to generate crowd counting maps as heatmaps. These maps provide the features of crowd distributions to capture the size of crowds. Specifically, a single frame is fed into a pre-trained VGG19~\citep{simonyan2015very} to produce a 2D crowd density matrix. Then, a crowd counting map is obtained from the density matrix. The model generates a set of crowd counting maps for all the odd-numbered frames. Subsequently, a pre-trained RAFT model~\citep{teed2020raft} is employed to produce optical flow maps. The model then uses the optical flow maps to capture motion features. Finally, the video frames, the crowd counting maps and the optical flow maps are concatenated to serve as the input for the swin transformer. Then a softmax layer receives the transformer's output to predict the probabilities of diverse crowd behaviours. 

The architecture of the swin transformer is shown in \cref{fig:swin} (a), including four stages. Suppose that the given video has the shape $T \times H \times W \times 3$, with the number of frames $T$, the frame height $H$, the frame width $W$, and 3 RGB channels. Subsequently, a 3D patch partition layer segments the video into a set of 3D patches with shape $2 \times 4 \times 4 \times 3$. Each 3D patch is flattened to be a token with a 96-dimensional feature. Therefore, there are $\frac{T}{2} \times \frac{H}{4} \times \frac{W}{4}$ tokens with dimension 96 as input for stage 1. The linear embedding layer converts the feature dimension of every token into an arbitrary dimension $C$. The video swin transformer block refines the features further. Next, three stages with the same structure are applied to all tokens to obtain more refined representations. The patch merging layer implements spatial downsampling by concatenating features of $2 \times 2$ neighbouring tokens and applies a linear layer to reduce the dimension of the merged features by half. Finally, the architecture of the video swin transformer block is shown in \cref{fig:swin} (b), which has two crucial components: 3D shifted window-based multi-head self-attention (3D SW-MSA) and MLP. The 3D SW-MSA is designed to replace the traditional multi-head self-attention for effective video recognition. More details of 3D SW-MSA can be found in \citet{qaraqe2024crowd,liu2022video}. There is a layer normalization (LN) before each component. Additionally, the block introduces the residual connection for 3D SW-MSA and MLP.                                             
\subsection{Conclusion}
\begin{table}[t]
    \centering
    \caption{The comparison of key techniques in crowd behaviour recognition.}
    \begin{tabular}{c|c|c|c}
    \hline      
    Technique  &  Accuracy    &  \,\,\, Data Requirement  \,\,\, &  \,\,\, Computational Cost  \,\,\,    \\
    \hline
    Traditional  & \,\,\, Low-Medium \,\,\, & Low & Low \\
    \hline
    CNN  & High  & High & Medium \\
    \hline
    RNN  & High  & High & Medium \\
    \hline 
    GNN & High & High & High \\
    \hline 
    \,\,\, Transformer \,\,\, & High& High & High \\
    \hline 
    \end{tabular}
    \label{tab:cmp_recog}
\end{table}

In this section, we initially outlined traditional machine learning methods from the holistic and the individual-based perspective. Subsequently, we provided a structured introduction to deep learning approaches, categorised according to their network architectures. To understand their relative strengths and weaknesses, we give a high-level qualitative comparison in Table \ref{tab:cmp_recog}. 

First, the introduction of deep learning has substantially improved the recognition accuracy compared with the traditional machine learning approaches. While comparison is qualitative, providing a fair quantitative comparison remains challenging due to the absence of unified experimental settings and datasets. Different methods often utilise diverse datasets, and even when using the same dataset, their experimental conditions vary. For instance, \citet{li2018deep} aimed to identify eight distinct crowd behaviour categories using the CUHK Crowd dataset \citep{shao2014scene}, whereas \citet{wei2020very} focused on classifying crowd behaviours as either heterogeneous or homogeneous. Furthermore, traditional machine learning methods relying on feature engineering typically require significantly less data and lower computational cost than deep learning approaches. In particular, GNNs and transformers generally demand more computational resources due to their complex architectures. In summary, as data availability and computational power continue to grow, deep learning methods represent both the current standard and the direction for future research in crowd behaviour recognition.

\section{Discussion}
\subsection{Effectiveness of Crowd Behaviour Analysis Methods}

This chapter reviewed two fundamental areas in crowd behaviour analysis: crowd behaviour prediction and recognition. Recent developments in deep learning have significantly advanced research within these domains. As a result, existing methods in crowd behaviour analysis demonstrate varied effectiveness depending on crowd types, environmental contexts, and specific applications.  

Based on the diverse behavioural patterns, crowds are often classified into low-density and high-density crowds \citep{zhao2018role}. Existing crowd analysis methods excel at modelling low-density crowd dynamics. For instance, various human trajectory prediction methods generally focus on low-density crowds, where individual tracking is feasible and trajectory data are available. Current trajectory prediction methods based on deep learning have shown strong performance in capturing low-density crowd dynamics \citep{yue2022human,shi2023trajectory}, achieving low errors in common evaluation metrics such as ADE and FDE. Particularly, physics-inspired deep learning methods also mitigate the problem of lacking explainability \citep{yue2022human}. In contrast, modelling high-density crowds dynamics is significantly more challenging because of complex interactions between individuals. Most recently, current research on crowd behaviour analysis has achieved breakthroughs in this area. Specifically, \citet{yue2024human} studied detailed physical interactions at the full-body level through a new human motion prediction task. They proposed an effective interaction model, resulting in accurate motion prediction. \citet{he2025learning} regarded high-density crowds as continuum active matters and proposed a new crowd material point method. The method can learn from in-the-wild videos, the most common available data for high-density crowds, and then effectively predict and simulate dense crowd movements. Nevertheless, the research on high-density crowds remains comparatively less developed, primarily due to data scarcity and modelling complexity.

Current crowd prediction and recognition research relies heavily on specific datasets \citep{borja2018short,schuetz2023review,zhang2025comprehensive}. Although they contain various scenes, \eg indoor/outdoor, these datasets can not cover all real-world situations. Existing methods have achieved excellent prediction and recognition performance on these benchmark datasets \citep{yuan2021agentformer,liu2022visual,qaraqe2024crowd}, demonstrating their effectiveness in diverse environmental contexts. However, these methods generally require further validation or fine-tuning on more real-world data when deployed in complex real-world situations \citep{velayutham2023analysis}. A key future direction is to further enrich the datasets, capturing crowds under diverse events, time, places, etc, and test the generalisability of these methods.

Crowd behaviour analysis methods play an important role in various practical applications such as emergency response, crowd management, and autonomous vehicle systems. For emergency response scenarios, crowd behaviour recognition approaches effectively identify abnormal or potentially dangerous situations, such as stampedes and sudden dispersals. Accurate detection facilitates early warning, improves situational awareness, and supports strategic evacuation planning \cite{wei2020very}. In crowd management applications, trajectory prediction methods can provide accurate future behaviour estimations, enabling proactive interventions to prevent overcrowding and bottlenecks \citep{tamaru2024enhancing}. Crowd behaviour prediction and recognition are essential components of autonomous vehicle systems, particularly within urban environments where autonomous vehicles must safely navigate around groups of pedestrians. Anticipation of pedestrians' behaviours for the next few seconds and identification of crowd behaviours provide plenty of behavioural cues, allowing the autonomous vehicles to make context-aware decisions \citep{rasouli2019autonomous,camara2020pedestrian}.

\subsection{Future Research}
After several decades of research in crowd behaviour analysis, we have observed a wave of new research in the past few years in the field, owning to the fast development of deep learning. Similar to how deep learning has influenced many fields outside of machine learning and AI, it is highly likely that the new research tools based on deep neural networks will become more and more deeply entrenched into the toolbox of crowd analysis. We expect to see the application of deep learning continue to expand more widely into crowd research, way beyond its current scope. There are several key challenges and changes which are likely to emerge in the near future.

The first change is the digital infrastructure development to enable more automated, comprehensive, and multi-modal data capture for crowd research. As mentioned at the beginning of this chapter, the two key elements in applying deep learning are data and tasks. Data, as the foundation of any data learning application, has so far appeared in the form of images and video in crowd research. This is due to the cameras are probably the most commonly used sensors everywhere. Other data based on smartphones, smart watches, GPS, \etc., have been actively explored but not at the same scale yet. We expect more types of sensors to be deployed in the near future providing abundant data, which are not only large in quantity but also of higher quality as well as in more modalities. This also has a strong synergy with the development of the Internet of Things and Smart City.

Along with the wider and deeper data collection on crowds, we expect to see new tasks being defined and solved using deep learning. This might shed light on some of the problems previously deemed to be extremely challenging. One instance is high-density crowds where crushes could happen. Currently, it is rare to see data of crushing and collapsing crowds which can be directly used by deep learning. The insufficient data size and the quality of the data are the main bottlenecks. With new sensors deployed and new data captured, it is possible for deep learning to do real-time analysis and prediction, to prevent crowd crushes. Actually, \citet{yue2024human} have already started in this direction, albeit the data was captured in a laboratory environment, not natural environments. In general, it is possible to re-solve most of the existing problems in crowd analysis using deep learning, when the data is ready and the old problem is reformulated into a form friendly to deep learning.

Another possible future direction is to systematically move from supervised learning tasks to unsupervised ones. Currently, no matter it is trajectory prediction or behaviour recognition, supervision signals need to be obtained either from human labelling or some other algorithms in the pre-processing stage. When data capture becomes more automated, fully automated pre-processing might be hard and human labelling will become prohibitively laborious. We expect to see more unsupervised learning methods being developed in deep learning, following their predecessors in statistical machine learning.

While deep learning will continue to expand its influence in crowd analysis, along with new data collected and new tasks defined, we expect that there are also difficulties in applying deep learning to some research topics in this field. This is because some data are hard to collect and some tasks are difficult to clearly define, in the way that deep learning model can be developed. Crowds are also studied from the perspective of social sciences, such as group dynamics and crowd psychology. While naturalist motion data collection is already challenging to collect, at least they are directly observable. In contrast, some data employed in social sciences are intrinsically hard to collect, \eg personality, affective states, \etc. Admittedly, it still might be feasible to collect such data for a small number of people. But we expect it is difficult to scale and automate this kind of data collection to meet the demanding requirements of deep neural networks. It will require AI researchers, social scientists, engineers and policymakers to jointly create new solutions, to enable us to leverage deep learning for solving these problems. 

\section{Conclusion}
In recent years, deep learning has significantly transformed the research in crowd behaviour analysis, offering powerful tools to predict and recognise crowd actions with remarkable accuracy. This chapter provides a comprehensive review of recent developments in two core tasks: crowd behaviour prediction and recognition. We briefly introduced traditional machine learning approaches and systematically reviewed deep learning methods. In particular, we presented representative deep learning methods in detail. Moreover, this chapter summarises and compares the key techniques in crowd behaviour prediction and recognition to understand their respective strengths and weaknesses. Finally, we discussed the effectiveness of crowd behaviour analysis methods across multiple aspects and provide practical suggestions for future research directions regarding this field. It is our expectation that this review will bring greater visibility to this rapidly developing field and inspires further research in the discussed areas.

\bibliographystyle{plainnat}
\bibliography{egbib} %

\end{document}